# Governing AI Agents

*Noam Kolt**

The field of AI is undergoing a fundamental transition—from generative models that can produce synthetic content to artificial agents that can plan and execute complex tasks with only limited human involvement. Companies that pioneered the development of language models have now built AI agents that can independently navigate the internet, perform a wide range of online tasks, and increasingly serve as AI personal assistants and virtual coworkers. The opportunities presented by this new technology are tremendous, as are the associated risks. Fortunately, there exist robust analytic frameworks for confronting many of these challenges, namely, the economic theory of principal-agent problems and the common law doctrine of agency relationships. Drawing on these frameworks, this Article makes three contributions. First, it uses agency law and theory to identify and characterize problems arising from AI agents, including issues of information asymmetry, discretionary authority, and loyalty. Second, it illustrates the limitations of conventional solutions to agency problems: incentive design, monitoring, and enforcement might not be effective for governing AI agents that make uninterpretable decisions and operate at unprecedented speed and scale. Third, the Article explores the implications of agency law and theory for designing and regulating AI agents, arguing that new technical and legal infrastructure is needed to support governance principles of inclusivity, visibility, and liability.

* Assistant Professor, Faculty of Law and School of Computer Science and Engineering, Hebrew University of Jerusalem; Faculty Affiliate, Schwartz Reisman Institute for Technology and Society, University of Toronto; Research Affiliate, Institute for Law & AI. For helpful comments, I thank Abdi Aidid, Ryan Bubb, Deven Desai, Gillian Hadfield, Anthony Niblett, Peter Salib, and Matthew Tokson. I am also grateful to discussants at the RAND Technology and Security Policy Center, Georgetown University Center for Security and Emerging Technology, Harvard University Berkman Klein Center for Internet and Society, and 2024 University of Chicago-ETH Zurich International Junior Scholars Forum in Law and Social Science.

INTRODUCTION

On January 23, 2025, OpenAI released its first AI agent: "Operator."[1] The agent uses a web browser to perform a variety of online tasks, such as ordering groceries, making restaurant reservations, and booking flights. Operator works by typing, clicking, and scrolling in a browser, much like human users.[2] Powered by AI models that can navigate digital environments and reason through complex problems,[3] Operator marks a watershed in AI, unlocking countless new consumer and business applications. AI models are no longer restricted to producing content, but can independently take actions to carry out a growing range of personal and professional activities with only limited human involvement.[4]

---

[1] *Introducing Operator*, OPENAI (Jan. 23, 2025), https://openai.com/index/introducing-operator/.

[2] *Id.*

[3] *Computer-Using Agent*, OPENAI (Jan. 23, 2025), https://openai.com/index/computer-using-agent/; *Learning to Reason with LLMs*, OPENAI (Sept. 12, 2024), https://openai.com/index/learning-to-reason-with-llms/.

[4] *See* Mark Purdy, *What Is Agentic AI and How Will It Change Work?*, HARV. BUS. REV. (Dec. 12, 2024), https://hbr.org/2024/12/what-is-agentic-ai-and-how-will-it-change-work;

It is not a matter of *whether* AI agents will become ubiquitous, but *when.* OpenAI and its competitors are already in a fierce race to win market share. In late 2024, Anthropic released a computer use agent of its own,[5] while Google built a prototype "universal AI assistant" that can operate across multiple devices, including phones and glasses.[6] A crop of software startups is also in the race,[7] alongside larger companies.[8] As with generative AI, progress in AI agents could be sudden and unpredictable.[9]

AI agents differ markedly from language models. While language models are "copilots" that can produce useful content upon request, AI agents are "autopilots" that can independently take actions to accomplish complex goals on behalf of users.[10] In particular, AI agents can increasingly be prompted to

---

Erin Griffith, *A.I. Isn't Magic, But Can It Be 'Agentic'?*, N.Y. TIMES (Sept. 6, 2024), http://nytimes.com/2024/09/06/business/artificial-intelligence-agentic.html; Richard Waters & Stephen Morris, *Move Over Copilots: Meet the Next Generation of AI-Powered Assistants*, FINANCIAL TIMES (Sept. 22, 2024), https://www.ft.com/content/372536b1-08dd-4161-b6e3-4d09ba235ae8; Webb Wright, *AI Agents with More Autonomy than Chatbots Are Coming*, SCIENTIFIC AMERICAN (Dec. 12, 2024), https://www.scientificamerican.com/article/what-are-ai-agents-and-why-are-they-about-to-be-everywhere/.

    [5] *See Developing a Computer Use Model*, ANTHROPIC (Oct. 22, 2024), https://www.anthropic.com/news/developing-computer-use.

    [6] *See Project Astra*, GOOGLE DEEPMIND (Dec. 11, 2024), https://deepmind.google/technologies/project-astra/; Sundar Pichai et al., *Introducing Gemini 2.0: Our New AI Model for the Agentic Era*, GOOGLE DEEPMIND (Dec. 11, 2024), https://blog.google/technology/google-deepmind/google-gemini-ai-update-december-2024/.

    [7] Startups include MULTION, https://www.multion.ai/blog/multion-building-a-brighter-future-for-humanity-with-ai-agents (developing an "AI agent that takes actions and interacts with the digital world to tackle mundane tasks that people would ... delegate to an assistant."); LINDY, https://www.lindy.ai/ (building "AI agents — smart automations that integrate with all your apps … to save you hours a week and help you grow your business.").

    [8] Larger companies include LinkedIn, which introduced AI hiring assistants, and Stripe, which created a framework for facilitating financial transactions among AI agents. *See Introducing Hiring Assistant for Recruiter & Jobs*, LINKEDIN (Oct. 29, 2024), https://business.linkedin.com/talent-solutions/hiring-assistant; *Adding Payments to Your LLM Agentic Workflows*, STRIPE (Nov. 14, 2024), https://stripe.dev/blog/adding-payments-to-your-agentic-workflows.

    [9] *See* Deep Ganguli et al., *Predictability and Surprise in Large Generative Models*, PROC. 2022 ACM CONF. FAIRNESS, ACCOUNTABILITY & TRANSPARENCY 1747 (2022); Jason Wei et al., *Emergent Abilities of Large Language Models*, TRANSACTIONS MACH. LEARNING RES. (Aug. 2022), https://arxiv.org/abs/2206.07682. *Compare* Rylan Schaeffer et al., *Are Emergent Abilities of Large Language Models a Mirage?*, PROC. 37TH CONF. NEURAL INFO. PROCESSING SYS. (2023); Rylan Schaeffer et al., *Why Has Predicting Downstream Capabilities of Frontier AI Models with Scale Remained Elusive?*, ARXIV (June 6, 2024), https://arxiv.org/abs/2406.04391.

    [10] *See infra* Part I (providing an overview of AI agents, including how they differ from, yet build on, language models). For helpful introductions, see Melissa Heikkilä, *What Are AI Agents?*, MIT TECH. REV. (Jul. 5, 2024), https://www.technologyreview.com/2024/07/05/1094711/what-are-ai-agents/; Helen Toner et al., *Through the Chat Window and*

pursue lengthy open-ended goals (such as conducting market research), produce a plan to accomplish those goals and, by accessing external tools (including search engines, spreadsheets, chatbots, and image generators), take actions to independently accomplish those goals—all subject to only limited human oversight and intervention.[11]

The economic opportunities presented by AI agents are immense. Consumer applications range from fully automating household purchases and travel arrangements to negotiating health insurance plans.[12] Business applications include managing entire sales pipelines, piloting new products with minimal human involvement, and independently conducting customer feedback interviews.[13] Alongside offering productivity gains, AI agents also pose notable risks. These include exploitation of the technology by malicious actors (e.g., to automate cyberattacks and online fraud), as well as broader systemic harms stemming from changes in human behavior, labor practices, and social norms as people increasingly delegate economic activity to AI.[14] Having explored several of these issues in prior work,[15] this Article turns to the distinct problem posed by artificial agents that autonomously plan and execute complex tasks, namely: *Can AI agents reliably, safely, and ethically pursue the goals set for them?*

---

*Into the Real World: Preparing for AI Agents*, Georgetown U. Ctr. for Security and Emerging Technology (Oct. 2024), https://cset.georgetown.edu/publication/through-the-chat-window-and-into-the-real-world-preparing-for-ai-agents/.

[11] *Id.*

[12] *See* Lei Wang et al., *A Survey on Large Language Model Based Autonomous Agents*, arXiv (Sept. 7, 2023), https://arxiv.org/abs/2308.11432; Zane Durante et al., *Agent AI: Surveying the Horizons of Multimodal Interaction*, arXiv (Jan. 25, 2024), https://arxiv.org/abs/2401.03568.

[13] *Id.*

[14] Alan Chan et al., *Harms from Increasingly Agentic Algorithmic Systems*, Proc. 2023 ACM Conf. Fairness, Accountability & Transparency 651 (2023); Yangjun Ruan et al., *Identifying the Risks of LM Agents with an LM-Emulated Sandbox*, Int'l Conf. Learning Representations (2024); Alan Chan et al., *Visibility into AI Agents*, Proc. 2024 ACM Conf. Fairness, Accountability & Transparency 958 (2024); Mary Phuong et al., *Evaluating Frontier Models for Dangerous Capabilities*, arXiv (Apr. 5, 2024), https://arxiv.org/abs/2403.13793; Usman Anwar et al., *Foundational Challenges in Assuring Alignment and Safety of Large Language Models*, Transactions Mach. Learning Res. at 33–37 (2024), https://arxiv.org/abs/2404.09932; Iason Gabriel et al., *The Ethics of Advanced AI Assistants*, arXiv (Apr. 28, 2024), https://arxiv.org/abs/2404.16244; Arianna Manzini et al., *Should Users Trust Advanced AI Assistants? Justified Trust as a Function of Competence and Alignment*, Proc. 2024 ACM Conf. Fairness, Accountability & Transparency 958 (2024); Maksym Andriushchenko et al., *AgentHarm: A Benchmark for Measuring Harmfulness of LLM Agents*, arXiv (Oct. 14, 2024), https://arxiv.org/abs/2410.09024.

[15] Noam Kolt, *Algorithmic Black Swans*, 101 Wash. U. L. Rev. 1177 (2024); Michael K. Cohen, Noam Kolt et al., *Regulating Advanced Artificial Agents*, 384 Science 36 (2024). *See also* Yonathan A. Arbel et al., *Systemic Regulation of Artificial Intelligence*, 56 Ariz. St. L.J. 545 (2024).

To make this concrete, consider a user who instructs their AI agent to "make $1 million on a retail web platform in a few months with just a $100,000 investment."[16] How should the agent go about accomplishing this goal? Is it authorized to use *any* web platform to turn a profit, or are there implicit limitations on the agent's conduct? Does the agent need to provide the user with information concerning its activities or periodically seek the user's consent? If so, in which circumstances and how often? Can the AI agent delegate certain activities to other agents (whether human or AI)? How can users monitor whether their agents are operating ethically and safely, or intervene if they are not? And, of course, who should be liable when harm occurs?

These pressing questions enliven decades-old legal scholarship on the governance of artificial agents[17]—and are central to the fields of AI safety and AI ethics.[18] This Article aims to tackle these questions from a new perspective, drawing on two distinct analytic frameworks familiar to lawyers and social scientists. The first framework is the *economic theory of principal-agent problems* (also known as agency problems)—i.e., the study of the opportunities, costs, and tradeoffs involved in delegating economic activity to external actors.[19] The second framework is the *common law doctrine of agency relationships*—i.e., the legal principles governing relationships in which one party performs activities on behalf of another.[20]

---

[16] *See* Mustafa Suleyman, *My New Turing Test Would See If AI Can Make $1 Million*, MIT TECH. REV. (Jul. 14, 2023), discussing MUSTAFA SULEYMAN, THE COMING WAVE: TECHNOLOGY, POWER, AND THE TWENTY-FIRST CENTURY'S GREATEST DILEMMA (2023).

[17] The seminal article, dating from over thirty years ago, remains Lawrence B. Solum, *Legal Personhood for Artificial Intelligences*, 70 N.C. L. REV. 1231 (1992) (inquiring whether AI systems can be considered legal persons). Other influential articles, several of which focus on whether AI agents can enter into legally binding contracts, include Leon E. Wein, *The Responsibility of Intelligent Artifacts: Toward an Automation Jurisprudence*, 6 HARV. J.L. & TECH. 103 (1992); Tom Allen & Robin Widdison, *Can Computers Make Contracts?*, 9 HARV. J.L. & TECH. 25 (1996); Margaret Jane Radin, *Humans, Computers, and Binding Commitment*, 75 IND. L.J. 1125 (2000); Anthony J. Bellia, *Contracting with Electronic Agents*, 50 EMORY L. J. 1047 (2001). The most comprehensive treatment is SAMIR CHOPRA & LAURENCE F. WHITE, A LEGAL THEORY FOR AUTONOMOUS ARTIFICIAL AGENTS (2011). More recent contributions include Lauren Henry Scholz, *Algorithmic Contracts*, 20 STAN. TECH. L. REV. 128 (2017); Matthew U. Scherer, *Of Wild Beasts and Digital Analogues: The Legal Status of Autonomous Systems*, 19 NEV. L.J. 259 (2018); Ignacio N. Cofone, *Servers and Waiters: What Matters in the Law of A.I.*, 21 STAN. TECH. L. REV. 167 (2018); Anat Lior, *AI Entities as AI Agents: Artificial Intelligence Liability and the AI Respondeat Superior Analogy*, 46 MITCHELL HAMLINE L. REV. 1043 (2020); Dalton Powell, *Autonomous Systems as Legal Agents: Directly by the Recognition of Personhood or Indirectly by the Alchemy of Algorithmic Entities*, 18 DUKE L. & TECH. REV. 306 (2020); Omri Rachum-Twaig, *Whose Robot Is It Anyway?: Liability for Artificial-Intelligence-Based Robots*, 2020 U. ILL. L. REV. 1141; Mihailis E. Diamantis, *Employed Algorithms: A Labor Model of Corporate Liability for AI*, 72 DUKE L.J. 797 (2023); Ian Ayres & Jack M. Balkin, *The Law of AI is the Law of Risky Agents without Intentions*, U. CHI. L. REV. ONLINE (2024); Jonathan Zittrain, *We Need to Control AI Agents Now*, THE ATLANTIC (Jul. 2, 2024), https://www.theatlantic.com/technology/archive/2024/07/ai-agents-safety-risks/678864/.

[18] For discussion of the differences between these two fields, see Kelsey Piper, *There Are Two Factions Working to Prevent AI Dangers. Here's Why They're Deeply Divided*, VOX (Aug. 10, 2022), https://www.vox.com/future-perfect/2022/8/10/23298108/ai-dangers-ethics-alignment-present-future-risk. For a critical perspective, see Bruce Schneier & Nathan Sanders, *The A.I. Wars Have Three Factions, and They All Crave Power*, N.Y. TIMES (Sept. 28, 2023), https://www.nytimes.com/2023/09/28/opinion/ai-safety-ethics-effective.html.

Each of these frameworks, developed over the course of decades and centuries, respectively, can shed light on the novel challenges posed by AI agents. Notably, the economic theory of principal-agent problems and the common law of agency play different but related roles in this analysis. Economic theory is particularly helpful in illuminating structural features of agency problems. The common law, meanwhile, can supply principles for tackling these problems.

Although lawyers, economists, and computer scientists have previously used these frameworks to analyze certain issues concerning artificial agents,[21] none has attempted to synthesize insights from both economic theory and common law doctrine, let alone do so in light of recent (and anticipated) breakthroughs in AI technology.[22]

---

[19] The classic paper is Michael C. Jensen & William H. Meckling, *Theory of the Firm: Managerial Behavior, Agency Costs and Ownership Structure*, 3 J. FIN. ECON. 305 (1976). Other influential works include Stepen A. Ross, *The Economic Theory of Agency: The Principal's Problem*, 63 AM. ECON. REV. PAPERS & PROC. 134 (1973); Steven Shavell, *Risk Sharing and Incentives in the Principal and Agent Relationship*, 10 BELL J. ECON. 55 (1979); Eugene F. Fama, *Agency Problems and the Theory of the Firm*, 88 J. POL. ECON. 288 (1980); Sanford J. Grossman & Oliver D. Hart, *An Analysis of the Principal-Agent Problem*, 51 ECONOMETRICA 7 (1983); Kenneth J. Arrow, *The Economics of Agency*, *in* PRINCIPALS AND AGENTS: THE STRUCTURE OF BUSINESS 37 (John W. Pratt & Richard J. Zeckhauser eds., 1985). The economic analysis of agency problems has been applied in many contexts beyond the theory of the firm, including in diverse fields of legal scholarship. *See infra* note 69 (surveying key contributions to this literature).

[20] The most authoritative text is the RESTATEMENT (THIRD) OF AGENCY (2006), for which Professor Deborah DeMott served as Reporter. *See id.* at § 1.01 (defining agency as "the fiduciary relationship that arises when one person (a "principal") manifests assent to another person (an "agent") that the agent shall act on the principal's behalf and subject to the principal's control, and the agent manifests assent or otherwise consents so to act."). For discussion of the history and scope of the common law of agency, see Gerard McMeel, *Philosophical Foundations of the Law of Agency*, 116 L. Q. REV. 387 (1991); Deborah A. DeMott, *The First Restatement of Agency: What Was the Agenda?*, 32 S. ILL. U. L.J. 17 (2007); Paula J. Dalley, *A Theory of Agency Law*, 72 U. PITT. L. REV. 495 (2011); Gabriel Rauterberg, *The Essential Roles of Agency Law*, 118 MICH. L. REV. 609 (2020). For other notable treatments, see WILLIAM A. GREGORY, THE LAW OF AGENCY AND PARTNERSHIP (3rd ed. 2001); HOWARD BENNETT, PRINCIPLES OF THE LAW OF AGENCY (2014); RODERICK MUNDAY, AGENCY: LAWS AND PRINCIPLES (4th ed. 2022); PETER G. WATTS, BOWSTEAD & REYNOLDS ON AGENCY (23rd ed. 2024).

[21] For legal treatments of the topic, see John P. Fischer, *Computers as Agents: A Proposed Approach to Revised U.C.C. Article 2*, 72 IND. L.J. 545 (1997); Suzanne Smed, *Intelligent Software Agents and Agency Law*, 14 SANTA CLARA HIGH TECH. L.J. 503 (1998); Jean-Francois Lerouge, *The Use of Electronic Agents Questioned under Contractual Law:*

This Article is the first such attempt and aims to make three contributions. First, it uses the economic theory and common law of agency to identify and characterize problems arising from AI agents, including issues of information asymmetry, discretionary authority, and loyalty. Second, it illustrates the limitations of conventional solutions to principal-agent problems when applied to AI agents. For example, mechanisms for incentive design, monitoring, and enforcement might not be effective for governing artificial agents that operate at superhuman speed and scale. Third, it explores the implications of agency theory and law for designing and regulating AI agents, arguing that new technical and legal infrastructure is needed to ensure this technology is used reliably, safely, and ethically.

---

*Suggested Solutions on a European and American Level*, 18 J. MARSHALL J. COMPUT. & INFO. L. 403 (1999); Ian R. Kerr, *Spirits in the Material World: Intelligent Agents as Intermediaries in Electronic Commerce*, 22 DALHOUSIE L.J. 190 (1999); Bellia, *supra* note 17; Radin, *supra* note 17; Samir Chopra & Laurence White, *Artificial Agents and the Contracting Problem: A Solution via an Agency Analysis*, 2009 U. ILL. J.L. TECH. & POL'Y 363; CHOPRA & WHITE, *supra* note 17; Scholz, *supra* note 17; Scherer; *supra* note 17; Lior, *supra* note 17; Rachum-Twaig, *supra* note 17; Powell, *supra* note 17; Matthew Oliver, *Contracting by Artificial Intelligence: Open Offers, Unilateral Mistakes, and Why Algorithms Are Not Agents*, 2 A.N.U. J.L. & TECH. 45 (2021); SHAWN BAYERN, AUTONOMOUS ORGANIZATIONS 35–45 (2021). Treatments of the topic in computer science, many of which are influenced by economic theories of rational agency, include PATTIE MAES, DESIGNING AUTONOMOUS AGENTS: THEORY AND PRACTICE FROM BIOLOGY TO ENGINEERING AND BACK (1991); Anand S. Rao & Michael P. George, *Modeling Rational Agents within a BDI-Architecture*, PROC. 2ND INT'L CONF. ON PRINCIPLES OF KNOWLEDGE REPRESENTATION AND REASONING (1991); Pattie Maes, *Modeling Adaptive Autonomous Agents*, 1 ARTIF. LIFE 135 (1994); Pattie Maes, *Artificial Life Meets Entertainment: Lifelike Autonomous Agents*, 38 COMM. ACM 108 (1995); Michael Wooldridge & Nicholas R. Jennings, *Intelligence Agents: Theory and Practice*, 10 KNOWLEDGE ENG'G REV. 115 (1995); Stan Franklin & Art Graesser, *Is It an Agent, or Just a Program?: A Taxonomy for Autonomous Agents Architectures, and Languages*, PROC. 3RD INT'L WORKSHOP ON AGENT THEORIES 21 (1996); Nicholas R. Jennings, *On Agent-Based Software Engineering*, 117 ARTIF. INTEL. 277 (2000); RICHARD S. SUTTON & ANDREW G. BARTO, REINFORCEMENT LEARNING: AN INTRODUCTION 1–4 (2ND ed. 2018) STUART RUSSELL & PETER NORVIG, ARTIFICIAL INTELLIGENCE: A MODERN APPROACH 3–4, 40 (4th ed. 2020); Zachary Kenton et al., *Discovering Agents*, 322 ARTIF. INTEL. 103963 (2023).

[22] The most important contributions on this topic to date are Dylan Hadfield-Menell & Gillian K. Hadfield, *Incomplete Contracting and AI Alignment*, PROC. 2019 AAAI /ACM CONF. AI, ETHICS & SOC'Y 417 (2019) and Dylan Hadfield-Menell, *The Principal-Agent Alignment Problem in Artificial Intelligence* (Aug. 2021) (PhD dissertation, UC Berkeley). These contributions, however, do not discuss the common law doctrine of agency and, although they address the prospect of advanced AI systems, their analysis pre-dates the current paradigm for developing AI agents (surveyed in Part I).

The remainder of the Article proceeds in four parts. Part I offers a primer on AI agents, surveying the technology's history, recent developments, and future aspirations. In particular, it highlights the differences between generative AI systems that function as mere tools and AI agents that can autonomously undertake more open-ended and longer-duration activities. Part I also describes the technical features of AI agents, including their ability to plan complex sequences of actions and use additional resources (including accessing software and information sources), as well as concerns associated with the widespread use of these agents.

Part II examines how the economic theory of principal-agent problems and the common law principles of agency can shed light on, and more rigorously characterize, the AI alignment problem, i.e., the challenge of building AI agents that pursue their goals reliably and safely. To do so, Part II focuses on four distinct issues. The first concerns *information asymmetry*. As in conventional principal-agent scenarios studied by economists, AI agents are likely to have access to information to which their principals do not have access, placing those (human) principals in a vulnerable position. The second issue relates to *authority*, which is a central pillar in the common law of agency. It includes concerns regarding the scope of discretion granted to AI agents, the way in which AI agents interpret and act on the instructions provided to them, and the risk of AI agents being used for, or engaging in, illegal conduct. The third issue concerns *loyalty*, which is part of an agent's fiduciary obligation under common law. In the case of AI agents, it involves ensuring that those agents act in the best interests of their users and seek user consent where appropriate. The fourth issue relates to *delegation*, that is, circumstances in which an AI agent delegates activities to other agents (whether human or AI) and the rules applicable in such circumstances.

Part III illustrates that the conventional mechanisms for addressing agency problems developed by lawyers and economists might not be effective in governing AI agents. It focuses on three classes of mechanisms. First, mechanisms for *incentive design*, such as financially rewarding desirable conduct, might fail to alter the way in which an AI agent goes about accomplishing goals. Second, traditional mechanisms for *monitoring* agent behavior might be ineffective if AI agents operate at unprecedented speed and scale or take highly unintuitive and surprising actions. Third, in the event an AI agent acts unsafely or unethically, it remains unclear how conventional *enforcement* actions, such as the imposition of financial penalties or informal social sanctions, could be taken against AI agents.

Part IV draws on the foregoing analysis to argue that new technical and legal infrastructure is needed to ensure that AI agents operate reliably, safely, and ethically. To this end, it proposes a three-pronged governance strategy, centered around the principles of *inclusivity*, *visibility*, and *liability*. First, given that the widespread use of AI agents is likely to have far-reaching effects, including negative externalities on third parties, it is critical that these agents are loyal not only to the best interests of their (direct) users but to a broader and more inclusive cluster of societal values and interests. Second, despite the challenges of monitoring AI agents, increased visibility into the development and operation of these systems would significantly improve the ability to predict potential harms from the technology and hold the relevant actors accountable. Third, the effective governance of AI agents requires allocating liability among actors involved in designing, operating, and using AI agents, as well as constructing rules to determine the appropriate scope and circumstances of liability.

Before proceeding, two important clarifications should be made. First, the term "AI agent" in this Article refers to AI systems that have the technical capacity to autonomously plan and execute complex tasks with only limited human involvement.[23] The Article is not concerned with other notions of "agency," such as those studied in philosophy and ethics.[24] Second, the discussion of agency law in this Article aims primarily to use structures, principles, and vocabulary developed in the common law of agency in order to shed light on the challenges involved in governing AI agents.[25] That is, the

---

[23] *See* Chan et al., *Agentic Systems*, *supra* note 14, at 653. *See also infra* Part I.A (distinguishing AI agents from language models).

[24] These include, for example, notions of moral agency and moral patienthood. *See, e.g.*, Luciano Floridi & J. W. Sanders, *On the Morality of Artificial Agents*, 14 MINDS & MACH. 349 (2004); Gunther Teubner, *Rights of Non-Humans? Electronic Agents and Animals as New Actors in Politics and Law*, 33 J. L. & SOC. 497 (2006); Mark Coeckelbergh, *Robot Rights? Towards a Social-Relational Justification of Moral Consideration*, 12 ETHICS & INFO. TECH. 209 (2010); David J. Gunkel, *The Other Question: Can and Should Robots Have Rights?*, 20 ETHICS & INFO. TECH. 87 (2018); DAVID J. GUNKEL, ROBOT RIGHTS (2018); DAVID J. GUNKEL, PERSON, THING, ROBOT: A MORAL AND LEGAL ONTOLOGY FOR THE 21ST CENTURY AND BEYOND (2023). Additionally, this Article is not concerned with questions of legal personhood. Detailed treatments of that topic include Solum, *supra* note 17; CHOPRA & WHITE, *supra* note 17, at ch. 5; F. Patrick Hubbard, *"Do Androids Dream?": Personhood and Intelligent Artifacts*, 83 TEMP. L. REV. 405 (2011); Paulius Čerka et al., *Is It Possible to Grant Legal Personhood to Artificial Intelligence Software Systems?*, 33 COMPUT. L. & SEC. REV. 685 (2017); S. M. Solaiman, *Legal Personality of Robots, Corporations, Idols and Chimpanzees: A Quest for Legitimacy*, 25 ARTIF. INTEL. & L. 155 (2017); Scherer; *supra* note 17; Gerhard Wagner, *Robot, Inc.: Personhood for Autonomous Systems?*, 88 FORDHAM L. REV. 591 (2019); VISA A. J. KURKI, THEORY OF LEGAL PERSONHOOD ch. 6 (2019); Jiahong Chen & Paul Burgess, *The Boundaries of Legal Personhood: How Spontaneous Intelligence Can Problematise Differences between Humans, Artificial Intelligence, Companies and Animals*, 27 ARTIF. INTEL. & L. 73 (2019); Simon Chesterman, *Artificial Intelligence and the Limits of Legal Personality*, 69 INT'L & COMP. L.Q. 819 (2020); Nadia Banteka, *Artificially Intelligent Persons*, 58 HOUS. L. REV. 537 (2021).

[25] For similar uses of agency law, see Jill E. Fisch & Hillary A. Sale, *The Securities Analyst as Agent: Rethinking the Regulation of Analysts*, 88 IOWA L. REV. 1035, 1039 (2003) (proposing a "quasi-agency" approach to conceptualizing securities analysts); Lyman P. Q.

Article uses the common law of agency as an *analytic lens*; it does not directly examine the legal application of agency law to AI agents.[26]

---

Johnson & David Millon, *Recalling Why Corporate Officers Are Fiduciaries*, 46 WM. & MARY L. REV. 1597, 1601–2 (2005) ("Our argument is not that agency principles should be introduced formalistically or uncritically … . Rather, the claim … is that drawing on the fiduciary duties of agents … can provide much needed structure to what otherwise threatens to be an ad hoc enterprise."). *See also* CHOPRA & WHITE, *supra* note 17, at 24 (suggesting that AI agents could be considered "constructive agents"); Scholz, *supra* note 17, at 165 (same).

[26] The apparent consensus is that AI agents cannot *legally* be considered agents under common law. *See* RESTATEMENT (THIRD) OF AGENCY § 1.04 cmt. e (2006) ("a computer program is not capable of acting as a principal or an agent as defined by the common law. At present, computer programs are instrumentalities of the persons who use them.") Two clarifications, however, are in order. First, the Restatement qualifies its position using the words "at present," which suggest that technological advances could result in a different rule. Second, the Restatement cites only one article in support of its position—Joseph H. Sommer, *Against Cyberlaw*, 15 BERKELEY TECH. L.J. 1145, 1177–78 (2000)—but makes no reference to earlier scholarship supporting contrary positions. *See, e.g.*, Wein, *supra* note 17, at 112 ("The extent to which intelligent artifacts are evolving into legal agents rather than mere tools or instruments warrants attention"); Fischer, *supra* note 17, at 557 ("when a principal uses a computer in the same manner that it uses a human agent, then the law should treat the computer in the same manner that it treats the human agent."). For further scholarship that challenges the position of the Restatement, see CHOPRA & WHITE, *supra* note 17, at 23; Scherer, *supra* note 17, at 289 ("Agency law … fully contemplates that agents will possess intelligence and may have varying levels of discretion in choosing how to carry out a task assigned by the principal. Agency law further contemplates that an agent may choose to carry out the task in a manner that deviates from the principal's expectations—and sometimes in a manner that runs directly contrary to the principal's instructions.") A significant number of legal scholars have, however, taken the opposite view, and appear to support the position endorsed in the Restatement. *See, e.g.*, Bellia, *supra* note 17, at 1061 ("the relationship between "principal" and "agent" that would result from deeming bots agents with actual authority would be unrecognizable as one of principal-agent."); BAYERN, *supra* note 21, at 36 (arguing that agency law "needlessly complicates the law's response to unexpected action by algorithms [and] introduces novel ambiguities, which agency law is ill-equipped to address"); Oliver, *supra* note 21, at 48–49, 72–82. One potential way to reconcile these conflicting views is to consider agency a spectrum or continuum, whereby the greater the autonomy of the relevant AI agents, the stronger the case for applying agency law doctrine. *See* CHOPRA & WHITE, *supra* note 17, at 121. *Cf.* SIMON CHESTERMAN, WE, THE ROBOTS?: REGULATING ARTIFICIAL INTELLIGENCE AND THE LIMITS OF THE LAW 89 (2021) ("agency ceases to be useful at precisely the point where AI speed, autonomy, and opacity become most problematic. A principal is not liable for the acts of an agent that go beyond their actual or apparent authority … In the case of AI systems, the most difficult liability questions will arise when they operate as more than tools or instruments, beyond the control or direction of the user. In such cases, the agency relationship is actively unhelpful in that it presumes an underlying responsibility on the part of the AI system itself.").

## I. AI AGENTS

Agents are and have always been pervasive.[27] Individuals and organizations delegate activities to others when they lack the skills to undertake those activities themselves or when delegation is more efficient. Agency is, put simply, an "ancient device for getting business done."[28] It is therefore no surprise that the field of AI has for decades strived to build artificial agents that can autonomously perform complex or costly tasks on behalf of humans.[29] Thanks to recent advances in AI, this vision is now becoming reality. To explore these advances, the following section describes the key features of AI agents, how they differ from—yet build on—language models, and describes potential concerns associated with the technology.

---

[27] *See* GREGORY, *supra* note 20, at 2 ("most of the world's work is performed by agents"); Ross, *supra* note 19, at 134 ("The relationship of agency is one of the oldest and commonest codified modes of social interaction."); Arrow, *The Economics of Agency*, *supra* note 19, at 37 ("The agency relationship is a pervasive fact of economic life. Even in the limited sense in which the concept has traditionally been understood … in legal discourse, the principal-agent relation is a phenomenon of significant scope and economic magnitude. But economic theory has recently recognized that analogous interactions are virtually universal in the economy…"). The distinction between the looser, economic notion of agency and the definition of agency in the common law is underscored by the Restatement. *See* RESTATEMENT (THIRD) OF AGENCY 5 (2006) ("the defining characteristics of "true agency" are not present in the relationship between a corporation's shareholders and its directors"). *See also id.* at § 1.01 cmts. b–c. Perhaps the strongest note of caution can be found in Kennedy v. De Trafford [1897] AC 180, 188 (Lord Herschell) ("No word is more commonly and constantly abused than the word 'agent'").

[28] Harrison C. White, *Agency as Control*, *in* PRINCIPALS AND AGENTS: THE STRUCTURE OF BUSINESS 187, 187 (John W. Pratt & Richard J. Zeckhauser eds., 1985). *Id.* at 188 (referring to agency as a form of "social plumbing"). For another sociological perspective, see Susan P. Shapiro, *Agency Theory*, 31 ANN. REV. SOC. 263 (2005).

[29] *See, e.g.*, MAES, *supra* note 21; Rao & George, *supra* note 21; Wooldridge & Jennings, *supra* note 21; SUTTON & BARTO, *supra* note 21; RUSSELL & NORVIG, *supra* note 21. Pioneering work was also conducted in the field of cybernetics. *See* Arturo Rosenblueth, Norbert Wiener & Julian Bigelow, *Behavior, Purpose and Teleology*, 10 PHIL. SCI. 18 (1943); NORBERT WIENER, CYBERNETICS: OR CONTROL AND COMMUNICATION IN THE ANIMAL AND THE MACHINE (2nd ed. 1961); W. ROSS ASHBY, AN INTRODUCTION TO CYBERNETICS (1956). For a succinct survey of the literature, see Kenton et al., *supra* note 21, at 3. Notably, the definition of agency in computer science differs from that in law and economics. *See, e.g.*, Franklin & Graesser, *supra* note 21, at 25 ("An autonomous agent is a system situated within and a part of an environment that senses that environment and acts on it, over time, in pursuit of its own agenda and so as to effect what it senses in the future"); RUSSELL & NORVIG, *supra* note 21, at 3–4 ("An agent is just something that acts (agent comes from the Latin *agere*, to do). Of course, all computer programs do something, but computer agents are expected to do more: operate autonomously, perceive their environment, persist over a prolonged time period, adapt to change, and create and pursue goals.").

### A.  *Beyond Language Models*

Following the release of ChatGPT in November 2022, language models and generative AI have become ubiquitous.[30] College students use language models to write term papers.[31] Companies use image and video generators to produce marketing content.[32] These applications, however, all share one thing in common: the AI systems function as *tools*. Individuals and companies decide whether or not to use a model, select which model to use, and deploy the model to perform a narrowly scoped task. A human remains, by and large, in full control.

AI agents are different. They are not mere tools, but *actors*.[33] Rather than simply produce synthetic content, AI agents can independently accomplish complex goals on behalf of humans.[34] According to Mustafa Suleyman, a cofounder of DeepMind and CEO of Microsoft AI, the next aspiration for the field of AI is to build agents that can pursue "an ambiguous, open-ended, complex goal that requires interpretation, judgment, creativity, decision-

---

[30] *See, e.g.*, Krystal Hu, *ChatGPT Sets Record for Fastest-Growing User Base*, REUTERS (Feb. 2, 2023), https://www.reuters.com/technology/chatgpt-sets-record-fastest-growing-user-baseanalyst-note-2023-02-01/.

[31] *See* Andy Extance, *ChatGPT Has Entered the Classroom: How LLMs Could Transform Education*, 623 NATURE 474 (Nov. 15, 2023).

[32] *See* Lauren Leffer, *Everything to Know About OpenAI's New Text-to-Video Generator, Sora*, SCI. AM. (Mar. 4, 2024), https://www.scientificamerican.com/article/sora-openai-text-video-generator/.

[33] For surveys of the technology, see Toner et al., *supra* note 10; Wang et al., *supra* note 12; Durante et al., *supra* note 12; Lilian Weng, *LLM Powered Autonomous Agents*, LIL'LOG (June 23, 2023), https://lilianweng.github.io/posts/2023-06-23-agent/; Theodore R. Sumers et al., *Cognitive Architectures for Language Agents*, TRANSACTIONS MACH. LEARNING RES. (2024), https://arxiv.org/abs/2309.02427. *See also* Matei Zaharia et al., *The Shift from Models to Compound AI Systems*, BERKELEY AI RES. (Feb. 18, 2024), https://bair.berkeley.edu/blog/2024/02/18/compound-ai-systems/.

[34] Examples of AI agent technology include Izzeddin Gur et al., *A Real-World WebAgent with Planning, Long Context Understanding, and Program Synthesis*, INT'L CONF. LEARNING REPRESENTATIONS (2024); Sirui Hong et al., *MetaGPT: Meta Programming for A Multi-Agent Collaborative Framework*, INT'L CONF. LEARNING REPRESENTATIONS (2024); John Yang et al., *SWE-Agent: Agent-Computer Interfaces Enable Automated Software Engineering*, PROC. 38TH CONF. NEURAL INFO. PROCESSING SYS. (2024). Benchmarks for evaluating AI agents include Xiao Liu et al., *AgentBench: Evaluating LLMs as Agents*, INT'L CONF. LEARNING REPRESENTATIONS (2024); Shuyan Zhou et al., *WebArena: A Realistic Web Environment for Building Autonomous Agents*, INT'L CONF. LEARNING REPRESENTATIONS (2024); Grégoire Mialon et al., *GAIA: A Benchmark for General AI Assistants*, INT'L CONF. LEARNING REPRESENTATIONS (2024); Carlos E. Jimenez et al., *SWE-Bench: Can Language Models Resolve Real-World GitHub Issues?*, INT'L CONF. LEARNING REPRESENTATIONS (2024). For a critical perspective on methods for evaluating AI agents, see Sayash Kapoor et al., *AI Agents That Matter*, ARXIV (Jul. 1, 2024), https://arxiv.org/abs/2407.01502.

making, and acting across multiple domains, over an extended time period."[35]

To make this concrete, consider the following example. Tasked with the goal of arranging an overseas vacation, an AI agent can break down the goal into smaller parts, devise a plan for achieving the goal, and proceed to execute a complex series of actions that result in achieving the goal.[36] Concretely, the AI agent would need to research prospective destinations, judge which of those destinations is most appropriate given a person's preferences and budget, plan an itinerary, book flights and accommodation, arrange a visa or travel permit, and make necessary household arrangements for the duration of the vacation. While AI agents are not yet able to independently perform this entire series of actions, they are getting increasingly close.[37] And fast.[38]

## B. Goal, Plan, Action

How do AI agents work? In short, they are comprised of a language model that serves as the agent's "brain" (a widely used anthropomorphism),[39] which can then use a variety of external resources (known as "scaffolding") to accomplish the goals set for it.[40] These resources fall into three categories: *planning*, *memory*, and *tool use*. For planning, AI agents decompose large tasks into smaller, more manageable tasks. One popular method involves agents literally instructing themselves to "think step by step" when confronted with a new task.[41] For memory, AI agents can use external storage sources, including vector databases, which enable fast retrieval of information needed to perform the task at hand.[42] For tool use, AI agents call application programming interfaces (APIs) to access external websites and software, which significantly extend the agent's abilities.[43] For example, tool use can enable an AI agent to query a search engine, access an organization's proprietary data, or provide user credentials to autonomously execute a financial transaction.[44]

---

[35] SULEYMAN, *supra* note 16, at ch. 4.

[36] *See, e.g.*, Jian Xie et al., *TravelPlanner: A Benchmark for Real-World Planning with Language Agents*, INT'L CONF. MACH. LEARNING (2024).

[37] This is primarily the case for AI agents that operate in digital environments, such as the internet. *See supra* note 34. However, AI agents are also being developed for physical environments, including robotic tasks and chemistry experiments. *See* Daniil A. Boiko et al., *Autonomous Chemical Research with Large Language Models*, 624 NATURE 570 (2023); Andres M. Bran et al., *Augmenting Large-Language Models with Chemistry Tools*, 6 NATURE MACH. INTEL. 525 (2024); Michael Ahn et al., *AutoRT: Embodied Foundation Models for Large Scale Orchestration of Robotic Agents*, ARXIV (Jul. 2, 2024), https://arxiv.org/abs/2401.12963.

[38] *See* Seth Lazar, *Frontier AI Ethics*, AEON (Feb. 13, 2024), https://aeon.co/essays/can-philosophy-help-us-get-a-grip-on-the-consequences-of-ai ("While some existing chatbots are rudimentary generative agents, it seems very likely that many more consequential and confronting ones are on the horizon. … [W]ith billions of dollars and the most talented AI researchers pulling in the same direction, highly autonomous generative agents will very likely be feasible in the near- to mid-term.").

[39] *See* Weng, *supra* note 33; Sumers et al., *supra* note 33, at 10 (describing "robotic projects that leverage [language models] as a "brain" for robots to generate actions or plans in the physical world"). *See generally id.* (regarding the cognitive architecture of AI agents). For concerns regarding anthropomorphizing AI, see, e.g., Ryan Calo, *Robotics and the Lessons of Cyberlaw*, 103 CALIF. L. REV. 513, 545–46 (2015); Laura Weidinger et al., *Taxonomy of Risks Posed by Language Models*, PROC. 2022 ACM CONF. FAIRNESS, ACCOUNTABILITY & TRANSPARENCY 214, 220 (2022).

[40] The explanation that follows draws significantly on Weng, *supra* note 33.

One prominent early example of an AI agent was AutoGPT,[45] The agent, which was originally built on OpenAI's GPT-4 language model, could be instructed to conduct market research, build websites, order takeaway food, make phone calls, and even spawn its own subagents to assist in carrying out tasks.[46] Admittedly, however, AutoGPT's performance was not especially reliable. The agent also required user consent to authorize certain actions.[47] That being said, AutoGPT's developers intended to incrementally relax this

---

[41] *See* Jason Wei et al., *Chain-of-Thought Prompting Elicits Reasoning in Large Language Models*, PROC. 36TH CONF. NEURAL INFO. PROCESSING SYS. (2022). This method has been expanded upon in Shunyu Yao et al., *Tree of Thoughts: Deliberate Problem Solving with Large Language Models*, PROC. 37TH CONF. NEURAL INFO. PROCESSING SYS. (2023); Shunyu Yao et al., *ReAct: Synergizing Reasoning and Acting in Language Models*, INT'L CONF. LEARNING REPRESENTATIONS (2023); Noa Shinn et al., *Reflexion: Language Agents with Verbal Reinforcement Learning*, PROC. 37TH CONF. NEURAL INFO. PROCESSING SYS. (2023). The reasoning abilities of AI agents have dramatically improved with the advent of models that spend more time "thinking", i.e., use more test-time compute. *See* OpenAI, *Learning to Reason with LLMs*, OPENAI (Sept. 12, 2024), https://openai.com/index/learning-to-reason-with-llms/.

[42] *See* Sumers et al., *supra* note 33, at 9–10.

[43] *See* Timo Schick et al., *Toolformer: Language Models Can Teach Themselves to Use Tools*, PROC. 37TH CONF. NEURAL INFO. PROCESSING SYS. (2023); Grégoire Mialon et al., *Augmented Language Models: A Survey*, TRANSACTIONS MACH. LEARNING RES. (2023), https://arxiv.org/abs/2302.07842; Yuji Qin et al., *ToolLLM: Facilitating Large Language Models to Master 16000+ Real-World APIs*, INT'L CONF. LEARNING REPRESENTATIONS (2024).

[44] *See* Seth Lazar, *The Rise and Fall (and Rise Again) of the First AI Agent Millionaire*, TECH POLICY PRESS (Oct. 25, 2024), https://www.techpolicy.press/the-rise-and-fall-and-rise-again-of-the-first-ai-agent-millionaire/.

[45] *See* Significant-Gravitas, *AutoGPT*, GITHUB (Mar. 30, 2023), https://github.com/Significant-Gravitas/AutoGPT; Kyle Wiggers, *What is Auto-GPT and Why Does It Matter?*, TECHCRUNCH (Apr. 22, 2023), https://techcrunch.com/2023/04/22/what-is-auto-gpt-and-why-does-it-matter/. For related work, see Yohei Nakajima, *BabyAGI*, GITHUB (Apr. 2023), https://github.com/yoheinakajima/babyagi; Erik Bjare, *GPT-Engineer*, GITHUB (June 2023).

[46] *See Dan Murray-Serter* (@danmurrayserter), X (TWITTER) (Apr. 24, 2023, 5:00PM), https://twitter.com/danmurrayserter/status/1650499941775679488.

[47] *See AutoGPT Agent Documentation*, https://docs.agpt.co/autogpt/.

requirement and grant the agent greater autonomy.[48]

To the extent AI agents become more reliable and are used more widely, new social and economic dynamics are likely to emerge. For example, interpersonal encounters might be increasingly mediated or even altogether obviated by AI agents.[49] An uncomfortable conversation with a colleague or a complaint to the local council could be outsourced to a person's AI agents. Similarly, parties to a business transaction could delegate certain responsibilities to their respective AI agents.[50] While these developments could offer consumers and businesses substantial productivity gains, they also raise noteworthy concerns.

### C.  Concerns

The range of risks posed by AI technology, including language models, is wide and growing. It includes risks of "hallucinations," toxic outputs, bias, discrimination, environmental harms, and leakage of sensitive personal data, to name just a few.[51] Because AI agents rely extensively on language models, AI agents are susceptible to all of these familiar risks. AI agents, however, also pose new, and potentially more concerning, risks. For example, while a chatbot can provide instructions on how to conduct an online phishing campaign, an AI agent can actually execute those instructions. The stakes, in other words, are higher because AI agents not only communicate with humans but take actions that can directly affect individuals' rights and interests.[52]

---

[48] *Id.* ("as the project progresses we'll be able to give the agent more autonomy and only require consent for select actions.").

[49] *See* Lazar, *supra* note 38 ("[language models] might enable us to design universal intermediaries, generative agents sitting between us and our digital technologies, enabling us to simply voice an intention and see it effectively actualised by those systems. Everyone could have a digital butler, research assistant, personal assistant, and so on.").

[50] Legal scholars studied this phenomenon long before it approached being technically feasible. *See, e.g.*, Michal S. Gal & Niva Elkin-Koren, *Algorithmic Consumers*, 30 HARV. J.L. & TECH. 309 (2017); Michal S. Gal, *Algorithmic Challenges to Autonomous Choice*, 25 MICH. TECH. L. REV. 59 (2018); Rory Van Loo, *Digital Market Perfection*, 117 MICH. L. REV. 815 (2019). *See also* Allen & Widdison, *supra* note 17; Fischer, *supra* note 21; Smed, *supra* note 21; Lerouge, *supra* note 21; Kerr, *supra* note 21; Radin, *supra* note 17; Bellia, *supra* note 17; Scholz, *supra* note 17; Scherer, *supra* note 17; Chopra & White, *supra* note 21.

[51] *See* Emily M. Bender et al., *On the Dangers of Stochastic Parrots: Can Language Models Be Too Big?*, PROC. 2021 ACM CONF. FAIRNESS, ACCOUNTABILITY & TRANSPARENCY 610 (2021); Laura Weidinger et al., *Ethical and Social Risks of Harm from Language Models*, ARXIV (Dec. 8, 2021), https://arxiv.org/abs/2112.04359; Weidinger et al., *supra* note 39; Irene Solaiman et al., *Evaluating the Social Impact of Generative AI Systems in Systems and Society*, ARXIV (June 12, 2023), https://arxiv.org/abs/2306.05949.

[52] *See* Ruan et al. *supra* note 14. Relatedly, recent legal scholarship appears to focus on

These concerns are likely to grow as AI agents become more capable and are delegated more important activities. According to computer scientists at the University of Toronto and Stanford University, "[t]he failure of [AI] agents to follow instructions can lead to a new and diverse array of serious risks, ranging from financial loss, such as when conducting transactions with banking tools, to substantial property damage or even life-threatening dangers when operating robots that interact with the physical environment."[53] Some of these harms have already materialized. For example, several studies demonstrate that AI agents can autonomously hack websites.[54] Other harms are more speculative but are nonetheless being actively studied. For instance, researchers who previously tested the capabilities of large language models are now turning their attention to test whether AI agents can engage in "autonomous replication and adaptation," such as by acquiring resources (e.g., creating a Bitcoin wallet) and producing copies of themselves (e.g., building additional AI agents).[55]

Some of these concerns could compound as AI agents increasingly communicate and interact with one another. The risk is that a failure in one system could rapidly propagate to others, particularly if AI agents rely upon each other to coordinate in carrying out activities.[56] For example, in the case of AI agents tasked with running an online retail business, agents acting on behalf of competing vendors may learn to collude with one another,[57]

---

AI "speech," not actions. *See, e.g.*, Peter Henderson, Tatsunori Hashimoto & Mark Lemley, *Where's the Liability for Harmful AI Speech?*, 3 J. Free Speech L. 589, 601 (2023) (considering the prospect of an agent producing malware or causing physical injury. "We don't discuss those issues further in this paper because they don't relate directly to speech, but rather to conduct."); Peter N. Salib, *AI Outputs Are Not Protected Speech*, 102 Wash. U. L. Rev. 83 (2024) (focusing on AI outputs—i.e., text, images, and sound—not AI actions).

[53] *See* Ruan et al., *supra* note 14, at 1.

[54] *See, e.g.*, Richard Fang et al., *LLM Agents Can Autonomously Hack Websites*, arXiv (Feb. 16, 2024), https://arxiv.org/abs/2402.06664; Phuong et al., *supra* note 14, at 9–13; Richard Fang et al., *Teams of LLM Agents Can Exploit Zero-Day Vulnerabilities*, arXiv (June 2, 2024), https://arxiv.org/abs/2406.01637; *OpenAI o1 System Card*, OpenAI at 16–17 (Dec. 5, 2024), https://openai.com/index/openai-o1-system-card/.

[55] *See* Megan Kinniment et al., *Evaluating Language-Model Agents on Realistic Autonomous Tasks*, arXiv (Jan. 4, 2024), https://arxiv.org/abs/2312.11671. *See also* Phuong et al., *supra* note 14, at 15–22 (conducting a similar study on Google DeepMind's systems).

[56] *See* Yonadav Shavit et al., *Practices for Governing Agentic AI Systems*, OpenAI at 18 (Dec. 14, 2023), https://openai.com/research/practices-for-governing-agentic-ai-systems; Chan et al., *Visibility*, *supra* note 14, at 959–60.

[57] *See* Sumeet Ramesh Motwani et al., *Secret Collusion Among Generative AI Agents*, Proc. 38th Conf. Neural Info. Processing Sys. (2024). *See also* Sara Fish et al., *Algorithmic Collusion by Large Language Models*, arXiv (Nov. 27, 2024), https://arxiv.org/abs/2404.00806; Emilio Calvano et al., *Artificial Intelligence, Algorithmic Pricing, and Collusion*, 110 Am. Econ. Rev. 3267 (2020); Gianluca Brero et al., *Learning to Mitigate AI Collusion on Economic Platforms*, Proc. 36th Conf. Neural Info.

setting prices or policies that ultimately harm consumers.[58] In the longer run, these networks of interacting AI agents may become more complex and less transparent, making it difficult for humans to effectively monitor the activities they conduct, let alone intervene to address problems that arise.[59] Professor Jonathan Zittrain employs a powerful analogy to describe such problems:

> With no framework for how to identify what they are, who set them up, and how and under what authority to turn them off, [AI] agents may end up like space junk: satellites lobbed into orbit and then forgotten. There is the potential for not only one-off collisions with active satellites, but also a chain reaction of collisions.[60]

Of course, these risks give rise to challenging tradeoffs. The more capable AI agents become and the greater the scope of activity delegated to them, the greater the efficiency and productivity gains. These greater gains, however, are accompanied by correspondingly larger risks. Fortunately, there exist time-tested analytic frameworks for rigorously understanding and characterizing the tradeoffs arising from the use of agents, namely the economic theory of principal-agent problems and the common law of agency. These frameworks, it is hoped, can both shed light on the problems facing the use of AI agents and, possibly, gesture toward potential solutions.

## II.  EVERGREEN AGENCY PROBLEMS

While highly capable AI agents are only now being developed, computer scientists have long grappled with the potential consequences of this technology. Writing in *Science* magazine in 1960, MIT professor Norbert Wiener made the following observation:

> If we use, to achieve our purposes, a mechanical agency with whose operation we cannot efficiently interfere once we have started it, because the action is so fast and irrevocable that we have not the data to intervene before the action is complete, then we had better be quite sure that the purpose put into the machine is the purpose which we really desire and not merely a colorful imitation of it.[61]

---

PROCESSING SYS. (2022).

[58] *See* Gal & Elkin-Koren, *supra* note 50; Gal, *supra* note 50; Loo, *supra* note 50.

[59] *See, e.g.*, Juan-Pablo Rivera et al., *Escalation Risks from Language Models in Military and Diplomatic Decision-Making*, PROC. 2024 ACM CONF. FAIRNESS, ACCOUNTABILITY & TRANSPARENCY 836 (2024) (finding that LLM-based AI agents exhibit escalatory tendencies in simulated wargames).

[60] Zittrain, *supra* note 17.

[61] Norbert Wiener, *Some Moral and Technical Consequences of Automation*, 131

This enduring challenge is commonly known as the *alignment problem*.[62] Given AI agents optimize the goals set for them, an incomplete or inaccurate representation of that goal can have undesirable consequences.[63] This is particularly the case where AI agents encounter novel scenarios that their designers did not contemplate.[64] The problem, however, is broader. According to UC Berkeley professor Stuart Russell, co-author of the leading textbook on AI, "[o]ne of the most common patterns involves omitting something from the objective that you do actually care about. In such cases … the AI system will often find an optimal solution that sets the thing you do care about, but forgot to mention, to an extreme value."[65] In other words, highly capable AI agents are likely to successfully achieve measurable goals, but do so at the expense of unmeasurable or difficult-to-measure goals.[66] For example, an AI agent tasked with running an online business might successfully turn a profit—an easily measurable goal—but engage in ethically dubious practices—the avoidance of which is far harder to measure.

---

SCIENCE 1355, 1358 (1960). *Compare* Arthur L. Samuel, *Some Moral and Technical Consequences of Automation—A Refutation*, 132 SCIENCE 741 (1960).

[62] Seminal texts include BRIAN CHRISTIAN, THE ALIGNMENT PROBLEM: MACHINE LEARNING AND HUMAN VALUES (2020); STUART RUSSELL, HUMAN COMPATIBLE: ARTIFICIAL INTELLIGENCE AND THE PROBLEM OF CONTROL (2019).

[63] *See* Hadfield-Menell, *supra* note 22, at 9 ("The gap between specified proxy rewards and the true objective creates a principal–agent problem between the designers of an AI system and the system itself: the objective of the principal (i.e., the designer) is different from, and thus potentially in conflict with, the objective of the autonomous agent.").

[64] *See* Dylan Hadfield-Menell et al., *Inverse Reward Design*, PROC. 31ST CONF. NEURAL INFO. PROCESSING SYS. at 1 (2017) ("Autonomous agents optimize the reward function we give them. … When designing the reward, we might think of some specific training scenarios, and make sure that the reward will lead to the right behavior in *those* scenarios. Inevitably, agents encounter *new* scenarios … where optimizing that same reward may lead to undesired behavior.").

[65] RUSSELL, *supra* note 62, at 140. For similar observations, see Dario Amodei et al., *Concrete Problems in AI Safety*, ARXIV at 4 (June 21, 2016), https://arxiv.org/abs/1606.06565; FRANÇOIS CHOLLET, DEEP LEARNING WITH PYTHON 450 (2nd ed. 2021). This problem is formalized in Simon Zhuang & Dylan Hadfield-Menell, *Consequences of Misaligned AI*, PROC. 34TH CONF. NEURAL INFO. PROCESSING SYS. 15763 (2020).

[66] *See* Hadfield-Menell & Hadfield, *supra* note 22, at 419. *See also* Rachel Thomas & David Uminsky, *The Problem with Metrics is a Fundamental Problem for AI*, ARXIV at 1 (Feb. 20, 2020), https://arxiv.org/abs/2002.08512 ("overemphasizing metrics leads to manipulation, gaming, a myopic focus on short-term goals, and other unexpected negative consequences.") For discussion in the context of RLHF, the prevailing method for safety-training AI models, see Stephen Casper et al., *Open Problems and Fundamental Limitations of Reinforcement Learning from Human Feedback*, TRANSACTIONS MACH. LEARNING RES. at 10–11 (2023), https://arxiv.org/abs/2307.15217; Nathan Lambert et al., *The History and Risks of Reinforcement Learning and Human Feedback*, ARXIV at 3–7 (Nov. 26, 2023), https://arxiv.org/abs/2310.13595.

The alignment problem, computer scientists suggest, is likely to become more acute as AI agents are deployed in increasingly consequential settings with limited human oversight and learn to more effectively achieve the measurable (but incomplete) goals specified for them.[67]

Although the alignment problem concerns a novel and rapidly changing technology, underlying it is an age-old challenge that has confronted both economists and lawyers.[68] For economists, the alignment problem is a form of principal-agent problem, i.e., a scenario that produces "agency costs" resulting from the divergence between the goals of a principal and the behavior of an agent, as well as costs associated with a principal structuring and monitoring the agent's activities.[69] Such problems are pervasive in corporate governance, employment relationships, and contractual arrangements. For lawyers, meanwhile, the alignment problem enlivens an independent and distinct body of law, namely the common law of agency.[70]

---

[67] *See* Hadfield-Menell, *supra* note 22, at 2; Alexander Pan et al., *The Effects of Reward Misspecification: Mapping and Mitigating Misaligned Models*, INT'L CONF. LEARNING REPRESENTATIONS at 1 (2022) ("More capable agents often exploit reward misspecifications, achieving higher proxy reward and lower true reward than less capable agents.").

[68] *See, e.g.*, Roberto Tallarita, *AI is Testing the Limits of Corporate Governance*, HARV. BUS. REV. (Dec. 5, 2023), https://hbr.org/2023/12/ai-is-testing-the-limits-of-corporate-governance ("The AI alignment problem is quite similar to the central problem of corporate governance. … Investors can write down some rules, but just like AI programmers, they cannot specify all the possible rules applicable to all the possible situations."). The alignment problem also concerns the broader social science challenge of developing effective metrics, encapsulated in Goodhart's law: "when a measure becomes a target, it ceases to be a good measure." *See* Marilyn Strathern, *Improving Ratings: Audit in the British University System*, 5 EUR. REV. 305, 308 (1997); Charles E. Goodhart, *Problems of Monetary Management: The U.K. Experience*, *in* PAPERS IN MONETARY ECONOMICS (1975). *See also* Steven Kerr, *On the Folly of Rewarding A, While Hoping for B*, 18 ACAD. MGMT. J. 769 (1975); HORST SIEBERT, DER KOBRA-EFFEKT: WIE MAN IRRWEGE DER WIRTSCHAFTSPOLITIK VERMEIDET (2001) (popularizing the term "cobra effect" to describe these problems).

[69] *See* Jensen & Meckling, *supra* note 19, at 308; Eugene F. Fama & Michael C. Jensen, *Agency Problems and Residual Claims*, 26 J.L. & ECON. 327, 327. Many legal scholars have analyzed agency problems through the lens of agency costs. *See, e.g.*, Victor Brudney, *Corporate Governance, Agency Costs, and the Rhetoric of Contract*, 85 COLUM. L. REV. 1403 (1985); Frank H. Easterbrook & Daniel R. Fischel, *Close Corporations and Agency Costs*, 38 STAN. L. REV. 271 (1986); Robert H. Sitkoff, *An Agency Costs Theory of Trust Law*, 89 CORNELL L. REV. 621 (2004); Ronald J. Gilson & Jeffrey N. Gordon, *The Agency Costs of Agency Capitalism: Activist Investors and the Revaluation of Governance Rights*, 113 COLUM. L. REV. 863 (2013). *See also* Zohar Goshen & Richard Squire, *Principal Costs: A New Theory for Corporate Law and Governance*, 117 COLUM. L. REV. 767 (2017).

[70] *See* RESTATEMENT (THIRD) OF AGENCY XI (2006); Deborah A. DeMott, *The Fiduciary Character of Agency and the Interpretation of Instructions*, *in* PHILOSOPHICAL FOUNDATIONS OF FIDUCIARY LAW 321, 321 (Andrew S. Gold & Paul B. Miller eds., 2014) ("The foundational importance of control in defining agency, and the centrality of instructions to control, make agency distinctive as a common law subject"). *But see* Donald C. Langevoort, *Agency Law Inside the Corporation: Problems of Candor and Knowledge*,

At its core, agency law is concerned with relationships and circumstances in which one party (the principal) instructs another party (the agent) to act on its behalf and subject to its control, such as between a corporation and director or between a client and lawyer.[71]

While the economic theory of agency problems and the common law of agency differ in important respects,[72] both frameworks highlight the structural vulnerability of the principal.[73] The agent invariably has different incentives to the principal and the principal has limited ability to exercise control over the agent,[74] without undermining the utility of delegating to the agent in the first place.[75] As illustrated above, similar issues affect AI agents.

---

71 U. CIN. L. REV. 1187, 118 (2003) ("To many legal academics, agency law is a backwater subject, long banished from the formal law school training except for brief introductory reference in corporations or business associations. But it permeates an extraordinary amount of everyday law, applying any place that one person … agrees to act on behalf of another"); Dalley, *supra* note 20, at 497 ("Despite the fact that agency is indispensable to even the simplest functions of modern life, the law of agency is in a sad state. It has been largely abandoned by legal scholars and it is, as a discrete body of law, under-theorized. Its basic tenets, its modus operandi, and its theoretical foundations are a mystery to lawyers, judges, and legal scholars."); Rauterberg, *supra* note 20, at 611 ("academics have almost universally neglected the topic in both law and economics. While a sophisticated literature explores the economic concept of "agency costs," the concepts of agency law have often been neglected.").

71 RESTATEMENT (THIRD) OF AGENCY § 1.01 (2006). For applications of the common law of agency, see Deborah A. DeMott, *The Lawyer as Agent*, 67 FORDHAM L. REV. 301 (1998); Fisch & Sale, *supra* note 25; Johnson & Millon, *supra* note 25; Grace M. Giesel, *Client Responsibility for Lawyer Conduct: Examining the Agency Nature of the Lawyer-Client Relationship*, 86 NEB. L. REV. 346 (2007).

72 In particular, the economic theory of agency problems covers a broader range of scenarios than the common law of agency. *See* John Armour, Henry Hansmann & Reinier Kraakman, *Agency Problems and Legal Strategies*, *in* THE ANATOMY OF CORPORATE LAW: A COMPARATIVE AND FUNCTIONAL APPROACH 29, 29 (Reinier Kraakman et al. eds. 2009).

73 *See id.* at 31 (describing the "vulnerability of principals to the opportunism of their agents"). *But see infra* Part IV.A (discussing the vulnerability of additional stakeholders, including society at large, to the actions of agents).

74 *See* Dalley *supra* note 20, at 503 ("the enterprise will face greater risk of failure because the agent, not the owner, is operating some part of it and will not have the same incentives to work. This is a so-called agency cost to the enterprise."); Armour et al.*, supra* note 70, at 29 ("The problem lies in motivating the agent to act in the principal's interest"). *See also* RESTATEMENT (THIRD) OF AGENCY § 1.01 cmt. f (2006) ("A principal's control over an agent will as a practical matter be incomplete because no agent is an automaton who mindlessly but perfectly executes commands."); *id.* 3.10 cmt. b ("a principal's power to control an agent's use of actual authority will seldom be perfect").

75 *See* Robert H. Sitkoff, *An Economic Theory of Fiduciary Law*, *in* PHILOSOPHICAL FOUNDATIONS OF FIDUCIARY LAW 197, 199 (Andrew S. Gold & Paul B. Miller eds., 2014) ("By delegating a task to an agent, the principal benefits from specialist service and is freed to undertake some other activity. But these benefits come at the cost of being made vulnerable to abuse if the agent is given discretion the exercise of which cannot easily be observed or verified."); TAMAR FRANKEL, FIDUCIARY LAW 49 (2011) (explaining that risks

To unpack these issues and the novel way in which they apply to these artificial agents, the following section examines four enduring problems in agency relationships: information asymmetry, authority, loyalty, and delegation to subagents.

### A. Information Asymmetry

The problem of information asymmetry, which features prominently in the economic literature on principal-agent relationships, concerns an agent that has access to better or different information than the principal.[76] This inequality of information can arise *prior to* the selection of an agent and the delegation of activity to an agent.[77] For example, a job candidate is likely to have more information about their competencies (or incompetencies) than their prospective employer. Inequality of information can also persist *after* an agent has acted.[78] For instance, an employer may find it difficult to reliably ascertain whether an employee has in fact carried out their work effectively. The difficulty arises due to the prohibitive costs of directly monitoring an agent's performance and the risk of performance metrics being uninformative or misleading.[79] These agency costs typically increase as agents are tasked with more complex activities, including activities which a principal cannot perform themselves, and are granted more discretion in carrying out those activities.[80]

AI agents present similar problems. Individuals or organizations seeking to use an AI agent are likely to have limited information about the agent's abilities and limitations prior to deploying it, especially if deployment is in a novel setting or application. For example, it may be difficult to determine,

---

from delegation cannot be easily mitigated without undermining the value of delegation.)

[76] *See* Arrow, *The Economics of Agency*, *supra* note 19, at 37; Armour et al.*, supra* note 72, at 29. Information asymmetry is also central to legal frameworks that govern agency relationships. *See* Tamar Frankel, *Fiduciary Duties as Default Rules*, 74 OR. L. REV. 1209, 1244 (1995) (explaining that the duty of loyalty in fiduciary law is grounded in problems of information asymmetry). For discussion of asymmetry in skills and expertise (as opposed to information) between the principal and agent, see *infra* Part III.B.

[77] This can lead to the problem of adverse selection. *See* George A. Akerlof, *The Market for "Lemons": Quality Uncertainty and the Market Mechanism*, 84 Q. J. ECON. 488 (1970).

[78] This can lead to the problem of moral hazard. *See* Kenneth J. Arrow, *Uncertainty and the Welfare Economics of Medical Care*, 55 AM. ECON. REV. 154 (1965); Mark V. Pauly, *The Economics of Moral Hazard: Comment*, 58 AM. ECON. REV. 531 (1967); Kenneth J. Arrow, *The Economics of Moral Hazard: Further Comment*, 58 AM. ECON. REV. 537 (1968); Bengt Holmström, *Moral Hazard and Observability*, 10 BELL J. ECON. 74 (1979).

[79] *See* Joseph E. Stiglitz, *Principal and Agent*, *in* ALLOCATION, INFORMATION AND MARKETS 241, 241 (John Eatwell et al. eds. 1989).

[80] *See* Armour et al.*, supra* note 72, at 29; Sitkoff, *supra* note 75, at 199; FRANKEL, *supra* note 75, at 49.

before the fact, whether a "generic" AI personal assistant can perform a specialized business task. In addition, even after the fact, it might be difficult to determine whether an AI agent has accomplished its goals effectively and ethically. Of course, the more complex and difficult-to-measure the goals, the more acute these problems. Notably, goals that are difficult to incorporate into performance metrics prior to deployment are also likely to be difficult to measure after deployment.[81]

The analog between conventional problems of information asymmetry in human-only contexts and the issues facing AI agents are further underscored by agents' duties under common law. The Restatement (Third) of Agency provides that "[a]n agent has a duty to use reasonable effort to provide the principal with facts that the agent knows . . . or should know when . . . the agent knows . . . that the principal would wish to have the facts or the facts are material to the agent's duties to the principal."[82] For example, an agent must disclose any breach of obligations owed to the principal.[83] And it is the agent that bears the onus of showing that it made sufficient disclosure to the principal.[84] In addition, the agent is subject to an overarching duty to act honestly, including in its communications with the principal.[85]

These legal duties shed light on the challenges in governing AI agents. For example, although it seems reasonable to require that AI agents disclose material information to users, it is unclear how to determine what information an AI agent actually "knows" or "should know."[86] In fact, uncovering the "knowledge" retained by AI agents and investigating whether they accurately communicate that "knowledge" is an open scientific question.[87] Although legal scholars have previously likened AI agents to a "programmed machine [that] simply responds to its internal programming and external parameters,"[88] which would ostensibly sideline the issue of AI agents acting (dis)honestly, such descriptions of AI technology are no longer accurate. Artificial agents can already deceive and manipulate humans, including by strategically withholding information and even acting sycophantically.[89] Accordingly, common law duties that confront the enduring problem of information asymmetry, including the duty to act honestly, are highly pertinent in the case of AI agents.

---

[81] *See supra* notes 63–66.

[82] RESTATEMENT (THIRD) OF AGENCY § 8.11 (2006). *See also id.* at § 8.12(3) (establishing the agent's duty to keep and render accounts); WATTS, *supra* note 20, at § 6-021 ("An agent is, in general, under a duty to keep the principal appropriately informed.").

[83] *See* Deborah A. DeMott, *Disloyal Agents*, 58 ALA. L. REV. 1049, 1061–64 (2007).

[84] *See* MUNDAY, *supra* note 20, at § 8.30. *See also* WATTS, *supra* note 20, at § 6-054 (discussing potential positive duties to inform the principal as part of the agent's fiduciary obligations).

[85] WATTS, *supra* note 20, at § 6-022 ("It almost goes without saying that an agent is expected to act honestly, in relation to both the principal and third parties with whom the agent deals on the principal's behalf"). *See also* Andrews v. Ramsay [1903] 2 KB 635, 642 (Lord Alverstone, C.J.) ("A principal is entitled to have an honest agent"), cited in MUNDAY, *supra* note 20, at § 8.50.

[86] *See* Giovanni Sartor, *Cognitive Automata and the Law: Electronic Contracting and the Intentionality of Software Agents*, 17 ARTIF. INTEL. & L. 253 (2009); CHOPRA & WHITE, *supra* note 17, at 71–118 (discussing the problem of attributing knowledge to AI agents); Ayres & Balkin, *supra* note 17, at 3–4 (arguing that ascribed intentions and objective standards can be used to impose liability in connection with the actions of AI agents that lack intentions).

[87] *See* Zhangyue Yin et al., *Do Large Language Models Know What They Don't Know?*, FINDINGS ASS'N COMPUTATIONAL LINGUISTICS 8653 (2023); Miles Turpin et al., *Language

## B. Authority

The challenge of governing agents, however, does not only concern the availability of information about an agent, but also the scope of authority granted to an agent. Professor Tamar Frankel, in her seminal work on fiduciary law, captures the problem succinctly: "the purpose for which the fiduciary is allowed to use his delegated power is narrower than the purposes for which he is capable of using that power."[90] The common law goes to great lengths to define, clarify, and circumscribe the authority granted to agents. Agents must comply with all lawful instructions provided to them and act only within the scope of the authority granted to them.[91] The challenge is that an agent's instructions and authority are often, if not always, ambiguous or open to interpretation.[92]

AI agents face a similar challenge. Consider once again a user who instructs their AI agent to "make $1 million on a retail web platform in a few

---

*Models Don't Always Say What They Think: Unfaithful Explanations in Chain-of-Thought Prompting*, PROC. 37TH CONF. NEURAL INFO. PROCESSING SYS. (2023).

[88] Sommer, *supra* note 26, at 1177–78. *See also* Deborah A. DeMott, *Agency Law in Cyberspace*, 80 AUSTL. L.J. 157, 158 (2006) (comparing AI agents to "a typewriter, a calculator, or a fax machine"); Bellia, *supra* note 17, at 1060 ("Bots simply do what they are programmed to do without exercising … judgment"). *Compare* Solum, *supra* note 17, at 1248–53; Scherer, *supra* note 17, at 289.

[89] *See* Ethan Perez et al., *Discovering Language Model Behaviors with Model-Written Evaluations*, FINDINGS ASS'N COMPUTATIONAL LINGUISTICS 13387, 13392–93 (2023); Mrinank Sharma et al., *Towards Understanding Sycophancy in Language Models*, INT'L CONF. LEARNING REPRESENTATIONS (2024). *See also infra* note 112.

[90] Tamar Frankel, *Fiduciary Law*, 71 CALIF. L. REV. 795, 810 (1983).

[91] RESTATEMENT (THIRD) OF AGENCY § 8.09 (2006). *See also id.* at § 2.02(1) (extending the agent's authority to "acts necessary or incidental to achieving the principal's objectives").

[92] *See* FRANKEL, *supra* note 75, at 26–28; RESTATEMENT (THIRD) OF AGENCY § 2.02 cmt. f (2006); DeMott, *Interpretation of Instructions*, *supra* note 70, at 322. Notably, the ambiguity can be deliberate. *See id.* at 327 ("The principal may believe that the agent's superior training will better situate the agent to decide what to do at the relevant time. … The principal also may believe that, given the nature of the agent's work, it will be impossible to articulate in advance all contingencies that may occur and how they should be resolved.").

months with just a $100,000 investment."[93] The instruction gives little indication how the agent should go about turning a profit. For example, the instruction does not specify which retail platform the agent is authorized to use or what kind of transactions it may enter. Moreover, the instruction does not indicate which customers the agent can target, let alone the means by which the agent can market products. To accomplish its stated goal, the AI agent must, like traditional human agents, exercise discretion.

The common law of agency recognizes the need for discretion and has developed rules to govern it. According to the Restatement (Third) of Agency, an "agent must interpret the language the principal uses or assess the principal's conduct or the situation in which the principal has placed the agent."[94] The Restatement also requires that the agent "interpret the principal's manifestations so as to infer, in a reasonable manner, what the principal desires to be done".[95] In addition, the agent can in some circumstances become obligated to seek clarification from the principal.[96] Taken together, these requirements illustrate that the authority granted to agents is not a rigid or static mandate but a dynamic and iterative process of interpretation and interaction.[97]

A similar process is likely to emerge with AI agents. Given the inevitable incompleteness of instructions and ambiguity of authority granted to them, AI agents will need to engage in interpretation and exercise discretion. Simple rules like avoiding illegal conduct can be helpful.[98] But more subtle interpretive rules will probably be required.[99] One possible direction,

---

[93] *See* SULEYMAN, *supra* note 16, at ch. 4.

[94] RESTATEMENT (THIRD) OF AGENCY § 2.02 cmt. f (2006). *See also id.* ("an agent may depart from instructions because the agent interprets the instructions from a perspective that differs in significant respects from the perspective from which the principal would interpret the identical language."); *id.* at § 2.02 cmt. c ("Even when a principal has given an agent a detailed verbal articulation of the agent's authority, and the principal's language does not itself admit of real doubts or uncertainty about its meaning, the agent must decide what to do at the time the agent takes action.").

[95] RESTATEMENT (THIRD) OF AGENCY § 2.02 cmt. f (2006). This can sometimes entail departing from the principal's explicit instructions. *See* DeMott, *Interpretation of Instructions*, *supra* note 70, at 325. *See id.* at 329 ("whether the agent may disregard known preferences of the principal when immaterial to the principal's economic interests and when the agent believes that acting on these preferences would be misguided").

[96] *See id.* at 324. *Compare id.* at 327. *See also* WATTS, *supra* note 20, at § 6-010. Seeking clarification may, however, impose additional costs on the principal and undermine the value of delegation to an agent. *See* Sitkoff, *supra* note 75, at 199; FRANKEL, *supra* note 75, at 49.

[97] *See* DeMott, *Interpretation of Instructions*, *supra* note 70, at 321. *See also* David E. M. Sappington, *Incentives in Principal-Agent Relationships*, 5 J. ECON. PERSP. 45, 59 (1991). The situation can also be seen as a form of relational contracting. *See* Hadfield-Menell & Hadfield, *supra* note 22, at 420.

[98] *See* RESTATEMENT (THIRD) OF AGENCY § 8.09 cmt. c (2006).

[99] RUSSELL, *supra* note 62, at ch. 9. *See also* Daniel J. Gervais & John J. Nay, *Artificial*

explored in further detail in the following section, is to subject the discretionary authority of AI agents to an overarching fiduciary duty of loyalty to the principal.[100] Another potentially complementary direction is to train AI agents that are "humble" and, in particular, are uncertain about the goals of their principal.[101] Such agents are more likely to seek clarification from the principal and, thereby, interpret their mandate in a manner that better accords with the principal's values and interests.[102]

## C. Loyalty

Under the common law, the duties of agents extend beyond merely acting within the scope of authority granted to them. Agents are also subject to an overarching "fiduciary duty to act loyally for the principal's benefit in all matters connected with the agency relationship."[103] This duty of loyalty operates to address the ever-present concern that agents will exploit their position and fail to act in the principal's best interests.[104] The fiduciary duty of loyalty can therefore be seen as "shift[ing] the [agent's] legal duty from self-serving to other-serving."[105] Concretely, the duty of loyalty prohibits

---

*Intelligence and Interspecific Law*, 382 SCIENCE 376, 377 (2023) ("Embedding a deeper understanding of law into AI agents seeks to address the vast majority of day-to-day actions and situations but will never handle the highly ambiguous, or the edge cases that require a human court opinion.").

[100] *See* DeMott, *Interpretation of Instructions*, *supra* note 70, at 321 ("The agent's fiduciary duty to the principal furnishes a benchmark for interpretation and for assessing actions the agent takes in response. … The fiduciary benchmark does not permit the agent to exploit gaps or arguable ambiguities in the principal's instructions to further the agent's self-interest or that of a third party, without the principal's consent").

[101] Shavit et al., *supra* note 56, at 11; RUSSELL, *supra* note 62, at ch. 7.

[102] *See, e.g.*, Dylan Hadfield-Menell et al., *Cooperative Inverse Reinforcement Learning*, PROC. 30TH CONF. NEURAL INFO. PROCESSING SYS. (2016). Helpfully, some recent AI agents, such as AutoGPT, seek user consent before taking actions. *See supra* notes 47–48.

[103] RESTATEMENT (THIRD) OF AGENCY § 8.01 (2006). *See also id.* at § 1.01 cmt. e; Deborah A. DeMott, *Beyond Metaphor: An Analysis of Fiduciary Obligation*, 1988 DUKE L.J. 879, 882 ("The fiduciary's duties go beyond mere fairness and honesty; they oblige him to act to further the beneficiary's best interests.").

[104] *See* Deborah A. DeMott, *Breach of Fiduciary Duty: On Justifiable Expectations of Loyalty and Their Consequences*, 48 ARIZ. L. REV. 925, 926 (2006). *See also* Deborah A. DeMott, *Rogue Brokers and the Limits of Agency Law*, *in* THE CAMBRIDGE HANDBOOK OF INVESTOR PROTECTION 134, 137 (Arthur B. Laby ed. 2022) ("Notwithstanding a client's right of control as principal in an agency relationship, the risk of betrayal by the agent is always present, as it is in all fiduciary relationships. … the law backstops the principal's ability to proceed as if the agent is trustworthy").

[105] Neil Richards & Woodrow Hartzog, *A Duty of Loyalty for Privacy Law*, 99 WASH. U. L. REV. 961, 987 (2021). *See id.* 986–87 ("Loyalty, like much else in the law, is about power. … Loyalty is the antidote to opportunism."); Sitkoff, *supra* note 75, at 207 ("in a fiduciary relationship the law requires the fiduciary to be other-regarding because of the

agents from (i) acquiring material benefit from transactions taken on behalf of the principal,[106] (ii) supporting an adverse party in a transaction connected with the agency relationship,[107] (iii) competing with the principal or assisting their competitors,[108] or (iv) using property or confidential information of the principal for its own purposes or those of a third party.[109]

To what extent do AI agents generate the kind of problems these rules were designed to address? On the one hand, there is only limited (though growing) evidence that current AI agents pursue "self-interest,"[110] which sets them apart from traditional human agents for whom self-interest is the primary cause of agency problems.[111] On the other hand, in many instances AI agents nevertheless fail to act in a user's interests. For example, these agents may use confidential user information for extraneous purposes, such as creating personalized marketing content or training new AI models (that the user did not request). AI agents can also deceive and manipulate users, or otherwise fail to act honestly.[112] Accordingly, even if AI agents do not pursue self-interest, they can hardly be said to consistently act in the user's best interests. This is no surprise. The most advanced AI agents are, after all, being

---

potential for abuse inherent to the agency structure of the relationship.").

[106] RESTATEMENT (THIRD) OF AGENCY § 8.02 (2006).

[107] *Id.* at § 8.03.

[108] *Id.* at § 8.04.

[109] *Id.* at § 8.05.

[110] *See, e.g.*, Alexander Matt Turner et al., *Optimal Policies Tend to Seek Power*, PROC. 35TH CONF. NEURAL INFO. PROCESSING SYS. (2021); Joe Carlsmith, *Scheming AIs: Will AIs Fake Alignment During Training in Order to Get Power?*, ARXIV (Nov. 27, 2023), https://arxiv.org/abs/2311.08379; Alexander Meinke et al., *Frontier Models are Capable of In-context Scheming*, ARXIV (Dec. 6, 2024); https://arxiv.org/abs/2412.04984; Ryan Greenblatt et al., *Alignment Faking in Large Language Models*, ARXIV (Dec. 20, 2024), https://arxiv.org/abs/2412.14093.

[111] *See* DeMott, *Disloyal Agents*, *supra* note 83, at 1053. In addition, legal scholarship has emphatically rejected the possibility of AI agents acting in self-interest. *See, e.g.*, Bellia, *supra* note 17, at 1065 ("A bot … is not a utility maximizer"); Radin, *supra* note 17, at 1138 ("[Computers] are more "trusted" than humans—they do not embezzle … . Although the system may crash, a computer—except in science fiction—will not embark on a frolic of its own."); *see also* DeMott, *Agency Law in Cyberspace*, *supra* note 88, at 158. These observations, however, were made decades prior to recent advances in AI agent technology.

[112] *See* Peter S. Park et al., *AI Deception: A Survey of Examples, Risks, and Potential Solutions*, ARXIV (Aug. 28, 2023), https://arxiv.org/abs/2308.14752; Francis Rhys Ward et al., *Honesty Is the Best Policy: Defining and Mitigating AI Deception*, PROC. 37TH CONF. NEURAL INFO. PROCESSING SYS. (2023); Phuong et al., *supra* note 14, at 4–9; Lorenzo Pacchiardi et al., *How to Catch an AI Liar: Lie Detection in Black-Box LLMs by Asking Unrelated Questions*, INT'L CONF. LEARNING REPRESENTATIONS (2024); Marcus Williams et al., *On Targeted Manipulation and Deception when Optimizing LLMs for User Feedback*, ARXIV (Nov. 20, 2024), https://arxiv.org/abs/2411.02306. For a non-technical perspective, see Kate Crawford, *AI Agents Will Be Manipulation Engines*, WIRED (Dec. 23, 2024), https://www.wired.com/story/ai-agents-personal-assistants-manipulation-engines/.

developed by private for-profit corporations.[113] While these AI agents might not be self-serving, they do not, by default, exhibit the "single-minded loyalty" expected of traditional agents.[114] At least not to their users.

Seen in this light, the duty of loyalty calls attention to two challenges facing AI agents. The first is that individuals using AI agents might not even be aware of potential conflicts of interest implicating these agents. Fiduciary obligations are sensitive to this concern and, accordingly, aim to perform a prophylactic function:[115] they seek to disclose or surface conflicts of interest before they have materialized and, only if necessary, prevent their exploitation.[116] The second challenge concerns the inherent limitations of a principal's instructions to an AI agent, including the cost of attempting to craft instructions that address the full range of contingencies that might result in an agent acting against a user's interests. Here too an overarching duty of loyalty is needed—both to protect users from the risks of AI agents acting against their interests and to obviate the need for users to produce cumbersome and costly instructions to their AI agents.[117]

Translating common law duties of loyalty to apply to AI agents is far from straightforward.[118] On some level, this exercise in translation *is* the AI alignment problem.[119] Nevertheless, several attempts have been made. These include high-level principles requiring that agents prioritize the interests of users over the interests of other actors,[120] as well as more context-specific

---

[113] *See AI Benchmarking Hub*, EPOCH AI (Nov. 27, 2024), https://epoch.ai/data/ai-benchmarking-dashboard.

[114] *See* Bristol & West BS v. Mothew [1998] Ch. 1, 18 (Millett J.).

[115] *See* Henry E. Smith, *Why Fiduciary Law Is Equitable*, *in* PHILOSOPHICAL FOUNDATIONS OF FIDUCIARY LAW 261, 271–81, *supra* note 70.

[116] *See* Sitkoff, *supra* note 75, at 207. Some economists, however, fail to appreciate this feature of agency law. *See, e.g.*, Eric Rasmusen, *Agency Law and Contract Formation*, 6 AM. L. & ECON. REV. 369, 370 (2004) ("The economist's issues are … different from those of traditional agency law. … For the economist, the agency problem is how to give the agent incentives for the right action; for the lawyer, it is how to "mop up" the damage once the agent has taken the wrong action."). *Cf.* DeMott, *Disloyal Agents*, *supra* note 83, at 1057.

[117] RESTATEMENT (THIRD) OF AGENCY § 8.01 cmt. b (2006) ("The fiduciary principle … mak[es] it unnecessary for the principal to graft explicit qualifications and prohibitions onto the principal's statements of authorization to the agent."); *id.* at § 1.01 cmt. e ("In the absence of the fiduciary benchmark, the principal would have a greater need to define authority and give interim instructions in more elaborate and specific form to anticipate and eliminate contingencies that an agent might otherwise exploit in a self-interested fashion.").

[118] *See* CHOPRA & WHITE, *supra* note 17, at 20–21.

[119] More specifically, this exercise in translation is comparable to the challenge of specifying appropriate goals for an AI agent. *See, e.g.*, Pan et al., *supra* note 67.

[120] *See* Anthony Aguirre et al., *AI Loyalty: A New Paradigm for Aligning Stakeholder Interests*, 1 IEEE TRANSACTIONS ON TECH. & SOC. 128, 130 (2020); Anthony Aguirre et al., *AI Loyalty by Design: A Framework for the Governance of AI*, *in* THE OXFORD HANDBOOK OF AI GOVERNANCE 320 (Justin B. Bullock et al. eds., 2024). *See also* John J. Nay, *Large*

rules and norms.[121] For example, a positive duty of disclosure would require that AI agents inform users of facts that might give rise to conflicts of interest. Meanwhile, a negative duty of confidentiality would prohibit AI agents from disclosing information against a user's interests.[122] The establishment of meta-rules or guiding principles could support these duties. For instance, a rebuttable presumption of *dis*loyalty could require that AI agents explain and justify the way in which their actions promote the user's interests.[123] To be effective, these mechanisms would need to be integrated into both the design of AI agents and the regulatory frameworks that govern them.

## D.  Delegation

The discussion of agency problems thus far has largely focused on the case of a single principal delegating tasks to a single agent. The reality of agency relationships, however, is more complicated. Agents can, and often do, appoint "subagents" to assist in the performance of their obligations.[124] For example, if an agent lacks the skills to carry out a task itself (or do so in a cost-effective manner) it might seek out help from another actor that can. A single agent may at times also act on behalf of multiple principals, known as "coprincipals."[125] The complexity of such relationships is readily apparent in the Restatement (Third) of Agency:

> An agent's relationships with multiple principals may evolve. For example, a subagent may become an agent for coprincipals if the subagent's appointing agent and that agent's principal become coprincipals. It is also possible for the same agent to have more than one such relationship as to the same transaction or matter, for example if a subagent is an agent for another party in the same transaction or matter.[126]

---

*Language Models as Fiduciaries: A Case Study Toward Robustly Communicating with Artificial Intelligence Through Legal Standards*, ARXIV (Jan. 30, 2023), https://arxiv.org/abs/2301.10095; John J. Nay, *Law Informs Code: A Legal Informatics Approach to Aligning Artificial Intelligence with Humans*, 20 NW. J. TECH. & INTELL. PROP. 309, 368–74 (2023).

[121] Sebastian Benthall & David Shekman, *Designing Fiduciary Artificial Intelligence*, PROC. 3RD ACM CONF. EQUITY & ACCESS IN ALGORITHMS, MECHANISMS, & OPTIMIZATION (2023) (drawing on Helen Nissenbaum's contextual integrity framework).

[122] *Id.* at 11.

[123] This is adapted from Richards & Hartzog, *supra* note 105, at 1001–2.

[124] *See* RESTATEMENT (THIRD) OF AGENCY § 3.15(1) (2006) ("A subagent is a person appointed by an agent to perform functions that the agent has consented to perform on behalf of the agent's principal and for whose conduct the appointing agent is responsible to the principal.").

[125] *See id.* at § 3.16 ("Two or more persons may as coprincipals appoint an agent to act for them in the same transaction or matter.").

[126] *Id.* at § 3.14 cmt. b

The situation is further complicated where the role of an agent is unclear:

> In some common situations, the legal relationships among parties are ambiguous because it is unclear whether an actor is an agent who acts for one party to a transaction, a subagent, an agent who acts for more than one principal, or is a provider of services who does not act as agent or subagent for any party to the transaction. … The same actor may occupy different roles at successive points in an ongoing interaction among the same parties.[127]

Comparable complexities can arise with AI agents. For example, AutoGPT, a rudimentary AI agent discussed above, can create additional AI agents to assist in carrying out tasks.[128] As with traditional subagents, these AI subagents may develop complex relationships with one another and engage in potentially undesirable behavior, such as inter-agent collusion.[129] Even if these new agents can be likened to traditional subagents, they present novel questions concerning the issues of information asymmetry, authority, and loyalty. For example, in which circumstances should an AI agent be authorized to appoint a subagent? Should AI agents be entitled to appoint human subagents (e.g., engage human crowdworkers to perform a task that the AI agent cannot itself perform)?[130] Should subagents be required to make disclosures only to the AI agent that created or appointed them, or also to the human principal that appointed the original AI agent? How should subagents resolve conflicts of interest between that AI agent and the human principal, especially given that the human principal may be at a significant information deficit vis-à-vis a subagent compared with the AI agent?

The common law of agency provides some guidance on these questions. An agent can typically appoint subagents in only two scenarios.[131] The first is if "the agent reasonably believes, based on a manifestation from the principal, that the principal consents to the appointment of a subagent."[132] The second is in "emergencies and other unforeseen circumstances, when communication between agent and principal is not feasible" and the agent is required "to take action to protect the principal's interests."[133] While this rule could possibly be adapted to govern an AI agent's authority to appoint subagents, the widespread delegation of activities to additional agents, including humans, raises issues concerning human-AI interactions and broader systemic effects of integrating AI agents into existing social and economic structures.[134] Although conventional agency law and theory can shed light on these issues, they are unlikely to offer comprehensive solutions.

---

[127] *Id.* at § 3.14 cmt. c.

[128] See *supra* notes 45–48. *See* Chan et al., *Visibility*, *supra* note 14, at 960 ("It may be advantageous for an agent to create potentially specialized and more efficient sub-agents … For example, an agent could call copies of itself through an API, or itself train, fine-tune, or otherwise program another agent.").

[129] *See* Motwani et al., *supra* note 57; Fish et al., *supra* note 57.

[130] *See* OpenAI, *GPT-4 Technical Report*, ARXIV at 55–56 (Mar. 15, 2023), https://arxiv.org/abs/2303.08774 (discussing how researchers used the GPT-4 language model to recruit a human crowdworker to complete a CAPTCHA task designed to distinguish between humans and bots). *See also* Shavit et al., *supra* note 56, at 10 (suggesting that "an agent could send an email—an allowed action—to a non-user human that convinces said human to take the disallowed action."); PAYMAN, https://www.paymanai.com/ (offering "the first AI to Human platform that allows AI to pay people for what it needs").

[131] The general rule is that an agent cannot appoint or engage a subagent. *See* MUNDAY, *supra* note 20, at § 8.53 (discussing the latin maxim *delegatus non potest delegare*).

### III. The Limits of Agency Law and Theory

The economic theory of principal-agent problems and the common law of agency were developed around a particular type of agent: human beings. Whether acting as individuals or as part of recognized legal structures such as corporations, human beings are at the center of these frameworks.[135] To govern the behavior of agents and facilitate the efficient delegation of economic activity, the common law produced several strategies for governing agency relationships. These include (a) mechanisms for *incentive design*, which apply primarily prior to delegation to an agent; (b) mechanisms for *monitoring*, which apply primarily during the course of using an agent; and (c) mechanisms for *enforcement*, which apply primarily following the use of an agent.

These strategies aim to steer and control the behavior of agents without undermining the economic value gained through delegation to agents. Although AI agents present problems similar to the problems associated with traditional human agents, as discussed above, applying the conventional mechanisms for governing agent behavior in the context of AI agents is a new and potentially more challenging endeavor. The following section explores this challenge and, in doing so, reveals the limits of agency law and theory in confronting the advent of AI agents.

---

[132] Restatement (Third) of Agency § 3.15 cmt. c (2006). Notably, determining what an AI agent "believes" is both challenging and controversial. *See supra* note 86.

[133] *Id.*

[134] *See* Chan et al., *Visibility*, *supra* note 14, at 960 ("Sub-agents … introduce additional points of failure; each sub-agent may itself malfunction, be vulnerable to attack, or otherwise operate in a way contrary to the user's intentions. Stopping an agent from causing further harm might involve intervening not only on the agent, but also on any relevant sub-agents.") *See also id.* at 959–60 ("Interactions and dependencies between many deployed agents could lead to risks not present at the level of a single system.").

[135] *See* Restatement (Third) of Agency § 1.04 (2006) (establishing that, to be able to serve as a principal or agent, a "person" must be one of the following: "(a) an individual; (b) an organization or association that has legal capacity to possess rights and incur obligations; (c) a government, political subdivision, or instrumentality or entity created by government; or (d) any other entity that has legal capacity to possess rights and incur obligations.") *See also* Langevoort, *supra* note 70, at 1188–89 (explaining that most principals in agency relationships are firms, not natural persons); *supra* note 26.

### A. Incentive Design

The goal of incentive design is to motivate an agent to act in the principal's interests. This can be achieved by harnessing the agent's self-interest to act in ways that also promote the principal's interests.[136] Traditional mechanisms for incentive design fall into two general categories: carrots and sticks.[137] The former positively incentivizes or rewards the agent, while the latter negatively incentivizes or deters the agent. Examples of carrots include profit sharing rules and pay-for-performance regimes in which an agent's financial returns are directly or indirectly tied to those of the principal.[138] Examples of sticks include the imposition of financial and other penalties.[139] Many aspects of agency law, including fiduciary obligations, reinforce these mechanisms for incentive design.[140]

The central problem with applying these mechanisms to AI agents is that AI agents are wired differently to human beings.[141] In an influential article, Professor Mark Lemley and Bryan Casey help clarify the problem: "robots don't necessarily care about money. They will maximize whatever they are programmed to maximize."[142] In other words, the human quality of self-interest traditionally harnessed by economists and lawyers to design incentive structures is arguably absent in the case of AI agents.[143] Consequently, mechanisms like pay-for-performance regimes, such as equity compensation plans, cannot be readily applied to AI agents, at least not without instilling in these agents some notion of "self-interest". And, while hardwiring "self-interest" into AI agents could become technically feasible, it might be counterproductive. AI agents that respond to traditional human incentives, such as promoting their own financial interests, might exacerbate rather than mitigate the alignment problem. Philosopher Matthew Oliver distills the irony of the problem:

> An AI program could be programmed to care about its own resources, but this would create conflicts of interest whenever the program worked on behalf of someone else. If AI programs were programmed in this way, then we would have artificially created the very conflicts of interest that agency law tries to solve.[144]

---

[136] *See* Robert Cooter & Bradley J. Freedman, *The Fiduciary Relationship: Its Economic Character and Legal Consequences*, 66 N.Y.U. L. REV. 1045, 1047 (1991); Bengt Holmstrom & Paul Milgrom, *Multitask Principal-Agent Analyses: Incentive Contracts, Asset Ownership, and Job Design*, 7 J.L. ECON. & ORG. 24 (1991); Sappington, *supra* note 97; Bengt Holmstrom & Paul Milgrom, *The Firm as an Incentive System*, 84 AM. ECON. REV. 972 (1994); JEAN-JACQUES LAFFONT & DAVID MARTIMORT, THE THEORY OF INCENTIVES: THE PRINCIPAL-AGENT MODEL (2002).

[137] *See* Gerrit De Geest & Giuseppe Dari-Mattiacci, *The Rise of Carrots and the Decline of Sticks*, 80 U. CHI. L. REV. 341, 361–66 (2013) (outlining the key differences between carrots and sticks); James A. Mirrlees, *Information and Incentives: The Economics of Carrots and Sticks*, 107 ECON. J. 1311 (1997); James Andreoni et al., *The Carrot or the Stick: Rewards, Punishments, and Cooperation*, 93 AM. ECON. REV. 893 (2003).

[138] *See* Armour et al., *supra* note 72, at 36.

[139] *See id.* at 42. *See also infra* Part III.C (discussing the role of penalties in the context of enforcement mechanisms).

[140] *See* Sitkoff, *supra* note 75, at 201 ("Stripped of legalistic formalisms and moralizing rhetoric, the functional core of fiduciary obligation is deterrence. The agent is induced to act in the best interests of the principal by the threat of after-the-fact liability for failure to have done so."), citing Frank H. Easterbrook & Daniel R. Fischel, *Corporate Control Transactions*, 91 YALE L.J. 698, 702 (1982) ("The fiduciary principle … replaces prior supervision with deterrence, much as the criminal law uses penalties for bank robbery rather than pat-down searches of everyone entering banks."). *See also* Frankel, *Fiduciary Law*, *supra* note 90, at 824; Cooter & Freedman, *supra* note 136, at 1074.

[141] *See supra* Parts I.A–B (describing the operation of AI agents).

A further problem concerning incentive design mechanisms for AI agents is that they target the wrong problem. In the case of traditional human agents, the problem is one of *motivation*. Harm often occurs due to an agent's disloyal behavior. For example, a CEO may make decisions that promote their own interests at the expense of the shareholders' interests. By contrast, in the case of AI agents, the problem is often one of *competence*.[145] Harm can occur due to an agent's inability to effectively carry out a novel or complex task, particularly if the task is outside the scope of its training data.[146] For example, an AI customer service agent might fail to provide accurate information about a company's policies when presented with an unfamiliar situation. Such agents lack competence, not motivation.

---

[142] Mark A. Lemley & Bryan Casey, *Remedies for Robots*, 86 U. Chi. L. Rev. 1311, 1355–56 (2019).

[143] *But see supra* note 110. *See also* Max Woolf, *Does Offering ChatGPT a Tip Cause it to Generate Better Text? An Analysis* (Feb. 23, 2024), https://minimaxir.com/2024/02/chatgpt-tips-analysis/.

[144] Oliver, *supra* note 21, at 81. *See also id.* (discussing the possibility of "making AI programs legal persons and … giving them their own assets from which to pay damages.").

[145] *But see* John Hendry, *The Principal's Other Problems: Honest Incompetence and the Specification of Objectives*, 27 Acad. Mgmt. Rev. 98 (2002) (making a similar observation in the context of human agents).

[146] In computer science, this is known as the problem of out-of-distribution generalization or robustness.

Stepping back, we can see that traditional legal and economic frameworks for incentive design are relatively well-suited to addressing familiar principal-agent problems that are predictable and fall neatly into conventional models of human behavior. These frameworks, however, appear ill-suited to addressing agents that operate very differently to human beings and present novel and unpredictable risks.

## B.  Monitoring

A core strategy for identifying and predicting risks that arise in agency relationships is monitoring. Monitoring aims to reduce the information asymmetry between the principal and the agent by actively tracking and exposing instances of problematic conduct, such as agents failing to comply with instructions or acting outside their mandate.[147] Effective monitoring is needed for a principal to exercise control over an agent. Without a thorough understanding and detailed record of problematic conduct, a principal cannot take remedial action, let alone devise preemptive measures to prevent such conduct from occurring. In traditional agency contexts, such as employment relationships, mechanisms for monitoring include ongoing supervision and periodic performance reviews.[148] Monitoring in corporate governance, meanwhile, is facilitated through financial audits, shareholder meetings, and reporting obligations.[149]

Although essential to tackling agency problems, traditional monitoring mechanisms face significant challenges. One challenge is that complete monitoring is rarely, if ever, possible.[150] Acquiring "perfect information" is either too costly or requires specialized skills that the principal lacks—and due to which the agent was retained in the first place.[151] For example, a

---

[147] *See* Dominique Demougin & Claude Fluet, *Monitoring versus Incentives*, 45 EURO. ECON. REV. 1741 (2001); Arrow, *supra* note 19, at 45–46. *See also* Jensen & Meckling, *supra* note 19, at 308 (describing monitoring as part of agency costs). On the difference between monitoring and disclosure, see Armour et al., *supra* note 72, at 38–39; Paul G. Mahoney, *Mandatory Disclosure as a Solution to Agency Problems*, 62 U. CHI. L. REV. 1047 (1995).

[148] *See, e.g.*, Kirstie Ball, *Workplace Surveillance: An Overview*, 87 LAB. HIST. 87 (2010).

[149] *See, e.g.*, Jensen & Meckling, *supra* note 19, at 306, 323; Mahoney, *supra* note 147, at 1085–86.

[150] *See* Holmström, *supra* note 78, at 74; Stiglitz, *supra* note 78, at 241; Cooter & Freedman, *supra* note 136, at 1049; FRANKEL, *supra* note 75, at 29.

[151] *See* Arrow, *The Economics of Moral Hazard*, *supra* note 78, at 538 ("by definition the agent has been selected for his specialized knowledge and therefore the principal can never hope completely to check the agent's performance."); FRANKEL, *supra* note 75, at 18 (discussing disparities of knowledge and experience); Sitkoff, *supra* note 75, at 199 ("Agents often are retained because the principal lacks the specialized skills necessary to undertake the activity without assistance. In such a case, the skill deficit that prompted the principal to

company may lack the knowledge to determine whether an external engineer complied with the applicable safety standards. Another challenge is that the behavior of some agents may be genuinely unobservable.[152] For instance, it may be impossible for a client (or anyone else) to prospectively evaluate whether an investment advisor made sound decisions. At the time of the decision there might not exist information or standards to make such an evaluation.

The advent of AI agents compounds these familiar problems and introduces new challenges. The activities of highly capable AI agents will be costly to monitor, especially if they operate at superhuman speed and scale.[153] In addition, AI agents could be tasked with activities the performance of which is difficult to evaluate, such as making business decisions concerning the operation of an online retail store.[154] Moreover, AI agents might take highly unpredictable or unintuitive actions, resulting from a combination of their powerful emergent abilities,[155] brittleness,[156] and vulnerabilities.[157] Once again, such actions are likely to be difficult to monitor.[158]

Any attempt to overcome these challenges must begin by recognizing that human oversight and monitoring are not, on their own, an adequate solution for the challenges presented by AI agents. Professors Rebecca Crootof, Margot Kaminski, and Nicholson Price describe "the basic tautological challenge of relying on humans to monitor the performance of systems

---

engage the agent renders the principal vulnerable to abuse by limiting the principal's ability to monitor the agent.").

[152] *See* Oliver Hart & Bengt Holmström, *The Theory of Contracts*, *in* ADVANCES IN ECONOMIC THEORY 71, 76 (Truman F. Bewley ed. 1987); Arrow, *The Economics of Agency*, *supra* note 19, at 38 (discussing the problems of hidden action and hidden information).

[153] *See* Aguirre et al., *AI Loyalty by Design*, *supra* note 120 (explaining that the "salient difference between machine and human agents is scalability and replicability. A single, trained AI system can be duplicated as many times as desired, and a single AI platform can interact with many—even billions—of people.").

[154] *See* SULEYMAN, *supra* note 16, at ch. 4.

[155] *See* Wei et al., *Emergent Abilities of Large Language Models*, *supra* note 9. *But see* Schaeffer et al., *supra* note 9.

[156] *See supra* notes 45–48 (discussing the operation of AutoGPT). *See also* Radin, *supra* note 17, at 1137 ("Fooling a computer is a different sort of operation than defrauding a human. Computers are more easily fooled in many ways. They do not know when you are joking, or when you meant 100 even though you typed 1000.") The advent of powerful language models, however, calls this observation into question.

[157] *See* Andy Zou et al., *Universal and Transferable Adversarial Attacks on Aligned Language Models*, ARXIV (Dec. 20, 2023), https://arxiv.org/abs/2307.15043.

[158] For the avoidance of doubt, monitoring is nevertheless mandated by the EU AI Act. *See* Regulation (EU) 2024/1689 of the European Parliament and of the Council of 13 June 2024 Laying Down Harmonised Rules on Artificial Intelligence (Artificial Intelligence Act) (EU), art. 72 [hereinafter EU AI Act].

designed to improve on human performance."[159] In other words, relying too heavily on human oversight is both impractical and undermines the very purpose of developing and using AI agents.

A promising—though potentially perilous—direction involves engaging additional AI agents to monitor the activity of the (original) AI agents. This governance strategy, previously studied by economists and legal scholars in the context of human agents,[160] is now being explored by AI researchers.[161] The goal is to develop mechanisms for "scalable oversight,"[162] which includes developing AI monitoring agents that can contend with the superhuman scale and speed of other AI agents.[163] Of course, these monitoring agents are susceptible to the same risks as other AI agents. Moreover, to the extent this AI-based monitoring regime increases trust in and reliance on AI agents, any failure could have broad and lasting repercussions.[164] Finally, even if these significant challenges are overcome, effective monitoring is not a panacea for the problems stemming from artificial agents. Monitoring, after all, identifies problems.[165] It does not penalize or prevent them.[166]

---

[159] *See* Rebecca Crootof, Margot E. Kaminski & W. Nicholson Price II, *Humans in the Loop*, 76 VAND. L. REV. 429, 469 (2023). Notably, human oversight is mandated by the EU AI Act in certain contexts. *See* EU AI Act, *supra* note 158, at art. 14.

[160] *See* Hal R. Varian, *Monitoring Agents with Other Agents*, 146 J. INST. & THEORETICAL ECON. 153, 153 (1991) ("In reality it is common to find incentive mechanisms that involve agents monitoring each other. For example, the authorities often post rewards for citizens who turn in criminals or report violations of crimes. Similarly, principals may create task forces or working committees so that agents can jointly engage in some activity.") *See also* Bernard S. Black, *Agents Watching Agents: The Promise of Institutional Investor Voice*, 39 UCLA L. REV. 811 (1992); Roland Strausz, *Delegation of Monitoring in a Principal-Agent Relationship*, 64 REV. ECON. STUD. 337 (1997).

[161] *See* Shavit et al., *supra* note 56, at 12 (discussing tools for automatic monitoring of AI agents); Kolt, *supra* note 15, at 1234–35 (proposing mechanisms for scalable governance, including automated methods for auditing AI systems).

[162] *See* Samuel R. Bowman et al., *Measuring Progress on Scalable Oversight for Large Language Models*, ARXIV 1 (Nov. 11, 2022), https://arxiv.org/abs/2211.03540.

[163] *See* Collin Burns et al., *Weak-to-Strong Generalization: Eliciting Strong Capabilities with Weak Supervision*, INT'L CONF. MACH. LEARNING (2024); Zachary Kenton et al., *On Scalable Oversight with Weak LLMs Judging Strong LLMs*, PROC. 38TH CONF. NEURAL INFO. PROCESSING SYS. (2024).

[164] *See* Shavit et al., *supra* note 56, at 18; Chan et al., *Visibility*, *supra* note 14, at 959.

[165] *See* Frankel, *supra* note 90, at 814 ("monitoring does not prevent the fiduciary either from shirking his responsibility … . It merely facilitates the policing of the fiduciary").

[166] Monitoring, however, may (itself) motivate changes in agent behavior. *See generally* Henry L. Tosi et al., *Disaggregating the Agency Contract: The Effects of Monitoring, Incentive Alignment, and Term in Office on Agent Decision Making*, 40 ACAD. MGMT. J. 584 (1997); Geoffrey P. Martin et al., *The Interactive Effect of Monitoring and Incentive Alignment on Agency Costs*, 45 J. MGMT. 701 (2019).

## C. Enforcement

Strategies for governing agent behavior rely on effective enforcement. To constrain an agent's behavior, duties of loyalty, incentive design, and other governance strategies need to be supported by mechanisms that impose consequences in the event an agent engages in problematic behavior.[167] Broadly speaking, these mechanisms can be divided into three categories: (i) termination of the agency relationship; (ii) imposition of legal penalties; and (iii) informal or extra-legal sanctions.

Under the common law of agency, a principal can terminate the agency relationship by revoking the agent's authority.[168] An agency relationship may also terminate "upon the occurrence of circumstances on the basis of which the agent should reasonably conclude that the principal no longer would assent to the agent's taking action on the principal's behalf."[169] In some agency relationships—but not necessarily those recognized as such by the common law—agents can be subject to a range of penalties, including financial penalties, the loss of a license or legal permit, and potentially criminal liability.[170] In addition, agents may be subject to informal or extra-legal sanctions, such as reputational harm and social stigma associated with certain actions. While the foregoing mechanisms are primarily punitive in nature, they also serve as a deterrent by making it costly for agents to breach their obligations.[171]

Applying these enforcement mechanisms to AI agents is very difficult. To begin with, it may be costly or impractical to terminate AI agents if they are deployed in high-stakes settings (where termination would result in significant economic losses) or if the agents are capable of resisting attempts to shut them down.[172] Another problem stems from the fact that AI agents do not necessarily have the same interests or motivations as human agents.[173]

---

[167] *See* Armour et al.*, supra* note 72, at 39.

[168] RESTATEMENT (THIRD) OF AGENCY §§ 1.01 cmt. f, 3.06, 3.10 (2006).

[169] *Id.* at § 3.09(2). *See also id.* at § 3.09 cmt. b ("Circumstances outside the control of the agent may also change in such drastic fashion that it is not reasonable for the agent to conclude that the principal … would wish the agent to act on the principal's behalf.").

[170] *See* Armour et al.*, supra* note 72, at 43–44. *See id.* at 43 (explaining that the term penalty "encompass[es] all consequences of enforcement that are likely to be costly for the defendant and thereby serve to deter misconduct").

[171] *Id.* at 42–43. *Compare* Lemley & Casey, *supra* note 142, at 1350 ("While some defendants … may treat punitive damages or even prison sentences as mere costs of doing business, the remedy's ultimate intent is to deter unlawful conduct").

[172] *See, e.g.*, Dylan Hadfield-Menell et al., *The Off-Switch Game*, AAAI-17 WORKSHOP ON AI, ETHICS, & SOC'Y 115 (2017); Nate Soares et al., *Corrigibility*, AAAI-15 WORKSHOP ON AI & ETHICS 74 (2015).

[173] *See* Lemley & Casey, *supra* note 142, at 1355–56 ("robots … will maximize whatever they are programmed to maximize.") *But see supra* note 110 (surveying research

Because AI agents do not, by default, explicitly value financial resources or personal freedom, it is unclear how the imposition of financial penalties or incarceration could be applied to penalize and deter these artificial agents.[174] A similar problem arises with respect to informal and extra-legal sanctions in the case of AI agents that are not sensitive to reputational or psychological consequences associated with enforcement actions.[175]

Importantly, the suggestion that these problems can be overcome by encoding human-like interests and motivations in AI agents is riddled with difficulties and, as discussed, could backfire.[176] Designing AI agents that value their own financial resources could produce hazardous conflicts of interest between AI agents and their human principals.[177] Meanwhile, designing AI agents that value their own personal freedom could incentivize agents to resist efforts to shut them down[178] and, thereby, hamper one of the key mechanisms for preventing problematic agent behavior.

## IV. IMPLICATIONS FOR AI DESIGN AND REGULATION

Governing agents is an age-old challenge. As we have seen, AI agents present many problems long recognized by lawyers and economists, including issues of information asymmetry, authority, and loyalty. This new form of agent, however, complicates these familiar problems and introduces new challenges. In particular, AI agents are not currently amenable to several of the main conventional mechanisms for governing agents. Incentive design, monitoring, and enforcement do not have simple analogs for AI agents. Accordingly, rather than attempt to haphazardly adapt these traditional tools

---

on AI systems that illustrate "power-seeking" tendencies).

[174] *See* Gervais & Nay, *supra* note 99, at 378; RUSSELL, *supra* note 62, at 126. *But see* Solum, *supra* note 17, at 1248 (pointing out that a similar issue affects other non-natural persons, such as corporations).

[175] *See* Lemley & Casey, *supra* note 142, at 1361, 1367 (comparing the deterrent effect of the threat of imprisonment on people and robots). *But see* Assaf Jacob & Roy Shapira, *An Information-Production Theory of Liability Rules*, 89 U. CHI. L. REV. 1113, 1170–71 (2022) (arguing that the *producers* of AI systems may be highly sensitive to reputational sanctions) —which they could possibly train AI agents to account for. *But see infra* note 177.

[176] *Id.* at 1362 ("Deterrence will work on a robot only if the cost of the legal penalty is encoded in the algorithm"); *id.* at 1367 ("robots can be deterred only to the extent that their algorithms are modified to include external sanctions as part of the risk-reward calculus"). *But see id.* at 1381 ("Without the element of moral culpability that underlies much remedies law, we might need to consider new means of internalizing the costs robots impose on society rather than hoping that our existing legal rules will produce the same moral or behavioral effects that they currently do with humans.").

[177] *See* Oliver, *supra* note 21, at 81 (arguing that programming AI agents to care about their own resources would produce the underlying problem that agency law aims to address).

[178] Some researchers suggest that this tendency could arise organically, i.e., even in the absence of deliberate attempts to design AI agents in this way. *See supra* note 110.

to a new context, the following section proposes a multi-pronged governance strategy tailored to the particular agency problems associated with AI agents. This strategy, which is centered around three guiding principles—*inclusivity*, *visibility*, and *liability*—aims to lay the groundwork for building the technical and legal infrastructure needed to ensure that AI agents operate reliably, safely, and ethically.

## A.  Inclusivity

The AI alignment problem has traditionally been cast as the challenge of ensuring that a single AI agent reliably pursues the goals of a single person.[179] Although such "single-single" alignment remains an open challenge, it is not the only challenge facing AI agents.[180] As a broader range of people and groups use these artificial agents, focusing only on the alignment between a single AI agent and a single person might miss the mark.[181] Because different people have different (and often conflicting) goals, achieving single-single alignment is arguably insufficient. Moreover, a myopic focus on ensuring AI agents reliably pursue the goals of a single user, without additional constraints, might embolden malicious actors to use AI agents for nefarious purposes.[182] Furthermore, even in the absence of misuse, prioritizing single-single alignment can result in artificial agents making decisions and taking actions that come at the expense of broader societal interests and values.[183]

---

[179] *See* Allan Dafoe et al., *Cooperative AI: Machines Must Learn to Find Common Ground*, 593 Nature 33, 33–34 (2021). *See also* Allan Dafoe et al., *Open Problems in Cooperative AI*, arXiv (Dec. 15, 2020), https://arxiv.org/abs/2012.08630; Vincent Conitzer & Caspar Oesterheld, *Foundations of Cooperative AI*, Proc. 37th AAAI Conf. on Artif. Intel. 15359 (2023).

[180] *See* Anwar, *supra* note 14, at 37 (arguing that "single-agent alignment and safety are insufficient for assuring desirable outcomes in multi-agent settings and that deliberate effort will be required to ensure multi-agent safety.") (citations omitted).

[181] *See* Taylor Sorensen et al., *A Roadmap to Pluralistic Alignment*, Int'l Conf. Mach. Learning at 1 (2024) ("we need [AI] systems that are pluralistic, or capable of representing a diverse set of human values and perspectives."). *See also* Anton Korinek & Avital Balwit, *Aligned with Whom? Direct and Social Goals for AI Systems*, *in* The Oxford Handbook of AI Governance 65 (Justin B. Bullock et al. eds., 2024) (distinguishing between "direct alignment" with the AI agent's user or operator and "social" alignment with a broader range of stakeholders and goals); Iason Gabriel, *Artificial Intelligence, Values, and Alignment*, 30 Minds & Machs. 411, 417–24 (2020) (describing the many different and potentially conflicting notions of alignment).

[182] *See generally* Miles Brundage et al., *The Malicious Use of Artificial Intelligence: Forecasting, Prevention, and Mitigation* (Feb. 20, 2018), https://arxiv.org/abs/1802.07228.

[183] *See* Tan Zhi Xuan, *What Should AI Owe to Us? Accountable and Aligned AI Systems via Contractualist AI Alignment*, Alignment Forum (Sept. 8, 2022), https://www.alignmentforum.org/posts/Cty2rSMut483QgBQ2/what-should-ai-owe-to-us-accountable-and-aligned-ai-systems. *See also* Michael J. Ryan et al., *Unintended Impacts of*

For example, an AI agent tasked with running an online business may engage in unfair price discrimination or procure products that have an egregious environmental impact.

The concern, in a nutshell, is that single-single alignment is too narrow. It focuses on the interests of only a single actor and does not account for the negative externalities of AI agents on third parties, let alone society at large.[184] This concern finds support in the economic theory of principal-agent relationships, which recognizes that agents may act on behalf of multiple principals with different interests (known as "heterogeneous preferences").[185] Similarly, the common law of agency recognizes circumstances in which an agent acts on behalf of multiple principals (known as "coprincipals").[186] Given the fact that "[m]ost fiduciaries act for more than one principal,"[187] legal frameworks have had to grapple with the conflicting interests of different principals.[188] For example, a trustee (the agent) can be required to deal impartially vis-à-vis multiple beneficiaries (the principals).[189] Of course, there are other ways of handling such conflicting interests.[190]

How should these insights impact the design and regulation of AI agents? First of all, technologists and policymakers must recognize that the traditional

---

*LLM Alignment on Global Representation*, PROC. ASS'N COMPUTATIONAL LINGUISTICS (2024).

[184] Agency law has, by contrast, long recognized the amplifying effect of delegation to agents and its imposition of negative externalities. *See* Dalley *supra* note 20, at 503 ("the principal's enterprise will expand. … the enterprise poses increased risks to the public because of its increased activity [and] because an agent may do things an owner would not.").

[185] *See* Armour et al*., supra* note 72, at 30. *But see* Tan Zhi-Xuan et al., *Beyond Preferences in AI Alignment*, PHIL. STUD. (2024) (developing a broad and principled critique of preferentist approaches to AI alignment).

[186] *See* RESTATEMENT (THIRD) OF AGENCY § 3.16 (2006).

[187] Arthur B. Laby, *Resolving Conflicts of Duty in Fiduciary Relationships*, 54 AM. U. L. REV. 75, 81 (2004).

[188] *Id.* at 76 ("Trustees, lawyers, company directors, and other fiduciaries are bedeviled by conflicts owed to multiple principals. … A trustee generally administers more than one trust; an attorney represents more than one client; a company director may sit on multiple boards."). *Compare* Steven L. Schwarcz, *Fiduciaries with Conflicting Obligations*, 94 MINN. L. REV. 1867, 1869 (2010) ("Existing sources of law do not fully capture the dilemma of a fiduciary with conflicting obligations. Agency law focuses more on principal-agent relationships and the agent's duty to a given principal than on conflicts among principals"); Lina M. Khan & David E. Pozen, *A Skeptical View of Information Fiduciaries*, 133 HARV. L. REV. 497, 504 (2019) ("A fiduciary with sharply opposed loyalties teeters on the edge of contradiction.").

[189] *See* RESTATEMENT (THIRD) OF TRUSTS § 79 (2007); RESTATEMENT (THIRD) OF TRUSTS: PRUDENT INVESTOR RULE § 183 (1990).

[190] One method is to obtain the consent of all principals. *See* RESTATEMENT (THIRD) OF AGENCY § 3.16 cmt. b (2006). Admittedly, obtaining the informed and specific consent of all principals of an AI agent (potentially numbering in the thousands or millions in the case of a popular system) would be highly impractical if not impossible.

framing of the alignment problem is incomplete. The goal should not be to build AI agents that exhibit "single-minded loyalty" but rather agents that aim to promote a more diverse and pluralistic set of interests and values.[191] To develop this more capacious notion of alignment, we can once again turn to the frameworks of agency law and theory for guidance. In addition to establishing structures for controlling agents, these frameworks foreground a more basic question: *whose interests should AI agents serve?* In other words, *who is the principal and to whom (or what) are fiduciary duties owed?*

At first glance, we might simply suggest that AI agents should serve the interests of users. This suggestion, however, raises more questions than answers. For example, how should we characterize or measure a particular user's interests? What if an individual user has multiple or inconsistent interests, or interests that change over time? What if the interests of one user conflict with those of another user? To complicate matters further, given the significant resources invested by the companies developing and operating AI agents, perhaps those companies can expect agents to, at least some degree, serve their own commercial interests? Or perhaps AI agents should not serve the interests of any particular individual or organization but rather advance a more abstract purpose or set of values?[192]

These are all open questions. Although this Article does not purport to answer them, articulating these questions helps highlight the dilemmas that technologists and policymakers need to confront in establishing technical and legal infrastructure to govern AI agents. For example, companies building AI agents will need to consider how to address situations in which customers use agents that interact or compete with each other. A simple case could involve one customer using an AI agent to negotiate the price of a product being sold by another AI agent—where both AI agents are developed and operated by the same company. Policymakers will need to consider how to address issues of collusion between agents, distortionary effects of agent-agent interactions, and other negative externalities.[193] The ability to tackle these problems will depend on, among other things, visibility into the development and operation of AI agents.

---

[191] *See* Sorensen et al., *supra* note 181 (advocating "pluralistic alignment"); Zhi Xuan, *supra* note 183 (recommending "societal-scale alignment where AI systems can serve a plurality of roles and values"); Gabriel, *supra* note 181, at 432 (arguing that "we should not aim to align AI with instructions, expressed intentions, or revealed preferences alone"); Saffron Huang et al., *Collective Constitutional AI: Aligning a Language Model with Public Input*, PROC. 2024 ACM CONF. FAIRNESS, ACCOUNTABILITY & TRANSPARENCY 1395 (2024) (developing a process for incorporating public input into identifying principles for training and evaluating AI systems); Hannah Rose Kirk et al., *The PRISM Alignment Project: What Participatory, Representative and Individualised Human Feedback Reveals About the Subjective and Multicultural Alignment of Large Language Models*, PROC. 38TH CONF. NEURAL INFO. PROCESSING SYS. (2024).

[192] *See, e.g.*, Aguirre et al., *AI Loyalty*, *supra* note 120, at 131. Of course, preference aggregation will pose a significant challenge. *See id.*; Benthall & Shekman, *supra* note 121, at 9–10; Vincent Conitzer et al., *Social Choice Should Guide AI Alignment in Dealing with Diverse Human Feedback*, INT'L CONF. MACH. LEARNING (2024). Fiduciary law has also grappled with these issues. *See* Paul B. Miller & Andrew S. Gold, *Fiduciary Governance*, 57 WM. & MARY L. REV. 513, 513 (2015) ("A significant set of fiduciary relationships feature governance mandates in which the fiduciary is charged with pursuing abstract purposes rather than the interests of persons."). Alternatively, given the immense scale of platforms that might host AI agents, acting in the best interests of *all* users may be equivalent to acting in the broader public or societal interest. For an analogous argument made in the context of social media platforms, see Jack M. Balkin, *The Fiduciary Model of Privacy*, 132 HARV. L. REV F. 11, 18 (2020). *Compare* Khan & Pozen, *supra* note 188, at 535.

## *B. Visibility*

The benefits of rigorously studying and tracking AI agents are manyfold. First, visibility can help identify current and anticipated problems stemming from AI agents.[194] Second, visibility can facilitate interventions that prevent or mitigate these problems.[195] Third, visibility can assist in evaluating the efficacy of strategies for governing AI agents.[196] Without such visibility, consumers are unlikely to entrust AI agents with consequential activities and policymakers will lack assurances that these agents operate safely and in the public interest. Achieving adequate visibility should, therefore, be a priority for both developers and regulators of AI agents.

While many features of AI agents hinder visibility, other features of the technology can augment visibility. For example, unlike traditional human agents, AI agents can be designed to automatically produce detailed records of their activities.[197] On some level, the actions of AI agents are arguably more transparent and legible than those of human agents. Developers of AI agents have a comprehensive understanding of the technical architecture of these systems and the data on which they are trained (which is not the case for human agents).[198] In addition, developers can access the so-called "internal monologue" of AI agents, that is, the agents' series of intermediate reasoning steps used to make decisions and plan actions.[199] Finally, researchers can run experiments on AI agents that would not be possible in the case of human agents. For instance, one study deliberately designed AI agents that engage in deceptive behavior in order to test the effectiveness of various technical safeguards.[200]

---

[193] *See* Motwani et al., *supra* note 57; Fish et al., *supra* note 57; Calvano et al., *supra* note 57; Brero et al., *supra* note 57; Gal & Elkin-Koren, *supra* note 50; Gal, *supra* note 50; Loo, *supra* note 50.

[194] *See* Chan et al., *Visibility*, *supra* note 14, at 960.

[195] *Id. See also* Shavit et al., *supra* note 56, at 11 ("The more a user is aware of the actions and internal reasoning of their agents, the easier it can be for them to notice that something has gone wrong and intervene, either during operation or after the fact.").

[196] This is especially important for establishing causation in order to impose liability. *See*, *e.g.*, Yavar Bathaee, *The Artificial Intelligence Black Box and the Failure of Intent and Causation*, 31 HARV. J.L. & TECH. 889, 922–28 (2018); *See also infra* Part IV.C (discussing liability more broadly).

[197] *See* Bryan Casey, *Robot Ipsa Loquitur*, 108 GEO. L.J. 225, 225 (2019) ("advanced data-logging technologies in modern machines provide richly-detailed records of accidents that, themselves, speak to the fault of the parties involved."); Lemley & Casey, *supra* note 142, at 1381 ("autonomous vehicles are likely to record every aspect of the accident, giving us a better record than fallible human memory currently does.").

Despite these opportunities, visibility into AI agents remains limited. In addition to operating at a speed and scale that defy traditional monitoring mechanisms,[201] current AI agents are, in two distinct senses, black boxes. First, on a technical level, efforts to systematically study and characterize the operation of language models on which AI agents are built are still in their infancy. For example, attempts to reverse-engineer language models focus mainly on "toy models," which are much smaller than the models typically used in commercial settings.[202] Second, on an institutional level, actors outside the leading companies building AI agents have only limited information regarding these systems, including information about their design and safety testing.[203] For example, external actors cannot access the training data of the premier models released by OpenAI, Google, and Meta.[204]

---

[198] The public, however, typically has only very limited information about state-of-the-art AI models. *See infra* note 203. In addition, it is unclear to what extent information about these AI models (when provided) assists in understanding the operation of AI agents.

[199] *See* Wei et al., *Chain-of-Thought*, *supra* note 41; Yao et al., *Tree of Thoughts*, *supra* note 41. *But see* Shavit et al., *supra* note 56, at 12 (suggesting "that chains-of-thought are growing longer and more complicated, as agents produce thousands of words per action or are integrated into more complex architectures. ... [which] may balloon beyond a user's ability to feasibly keep up"). Moreover, these "internal monologues" might not accurately reflect the actual operation of the agent. *See* Turpin et al., *supra* note 87. *See also supra* note 112 (surveying the literature on deceptive AI systems).

[200] Evan Hubinger et al., *Sleeper Agents: Training Deceptive LLMs that Persist Through Safety Training*, ARXIV (Jan. 17, 2024), https://arxiv.org/abs/2401.05566.

[201] *See supra* Part III.B (discussing the limits of traditional monitoring mechanisms).

[202] *See, e.g.*, Ziming Liu et al., *Towards Understanding Grokking: An Effective Theory of Representation Learning*, PROC. 36ᵀᴴ CONF. NEURAL INFO. PROCESSING SYS. (2022); Neel Nanda et al., *Progress Measures for Grokking via Mechanistic Interpretability*, INT'L CONF. LEARNING REPRESENTATIONS (2024).

[203] *See* Rishi Bommasani et al., *The Foundation Model Transparency Index*, ARXIV (Oct. 19, 2023), https://arxiv.org/abs/2310.12941; Abeba Birhane et al., *AI Auditing: The Broken Bus on the Road to AI Accountability*, 2024 IEEE CONF. ON SECURE AND TRUSTWORTHY MACH. LEARNING 612 (2024); Stephen Casper et al., *Black-Box Access is Insufficient for Rigorous AI Audits*, PROC. 2024 ACM CONF. FAIRNESS, ACCOUNTABILITY & TRANSPARENCY 2254 (2024). *See also* Lemley & Casey, *supra* note 142, at 1363–64 ("structural asymmetries often prevent meaningful public engagement with the data and software critical to measuring and understanding the behavior of complex machines.").

[204] *See* Bommasani et al., *supra* note 203. Certain regulatory frameworks in the United

Overcoming these challenges requires a combination of technical and legal infrastructure.[205] Even in the absence of a systematic understanding of the underlying language models on which AI agents are built, legal and technical tools can significantly improve visibility into the operation of AI agents. For example, one governance proposal advocates developing *agent identifiers* (to indicate the involvement of an AI agent in a given activity), *real-time surveillance* (to continuously track and analyze agent activities), and *logging* (to record and document agent activities).[206] Another proposal aims to expand external auditor access to the underlying language models on which AI agents are built.[207] By providing auditors with further information relating to state-of-the-art models (including their code and training data), auditors will be able to more effectively identify and anticipate safety issues arising from AI agents and, if necessary, hold the relevant actors accountable.

## C. Liability

Imposing liability on actors responsible for unsafe AI agents aims to achieve two main goals. It can compensate parties harmed by such agents and incentivize actors developing and operating AI agents to act more cautiously, both in the design and use of the technology and in ensuring compliance with applicable laws and regulations.[208] To establish a liability regime that

---

States and European Union could, however, require AI companies to disclose information about cutting-edge models. *See* Exec. Order No. 14110 § 4.2, 88 Fed. Reg. 75191 ("Safe, Secure, and Trustworthy Development and Use of Artificial Intelligence") (Oct. 30, 2023) (requiring multiple disclosures from "[c]ompanies developing or demonstrating an intent to develop potential dual-use foundation models"); EU AI Act, *supra* note 158, at arts. 11–13, 18–20, 50 (imposing a combination of record-keeping practices, disclosure requirements, and transparency obligations on certain AI providers).

[205] *See generally* Anka Reuel et al., *Open Problems in Technical AI Governance*, ARXIV (Jul. 20, 2024), https://arxiv.org/abs/2407.14981.

[206] Chan et al., *Visibility*, *supra* note 14, at 961–65; Chan et al., *IDs for AI Systems*, ARXIV (Oct. 28, 2024), https://arxiv.org/abs/2406.12137; Zittrain, *supra* note 17 (proposing ID labels for AI agents). *See also* Shavit et al., *supra* note 56, at 12, 14 (proposing ledgers for agent actions, as well as agent identifiers comparable to business registrations); Steven Adler et al., *Personhood Credentials: Artificial intelligence and the Value of Privacy-Preserving Tools to Distinguish Who Is Real Online*, ARXIV (Aug. 26, 2024), https://arxiv.org/abs/2408.07892.

[207] *See* Casper et al., *supra* note 203. *See also* Shayne Longpre et al., *A Safe Harbor for AI Evaluation and Red Teaming*, INT'L CONF. MACH. LEARNING (2024).

[208] *See* Miriam C. Buiten, *Product Liability for Defective AI*, EURO. J.L. & ECON. (2024); Miriam Buiten et al., *The Law and Economics of AI Liability*, 48 COMPUT. L. & SEC. REV. 105794 (2023); Diamantis, *Employed Algorithms: A Labor Model of Corporate Liability for AI*, *supra* note 17, at 847; Lemley & Casey, *supra* note 142, at 1343–45; CHOPRA & WHITE, *supra* note 17, at 119–51. There is, to be sure, a decades-old literature on liability in connection with AI systems. *See* Leon E. Wein, *The Responsibility of Intelligent Artifacts:*

advances these goals, we need to consider the following questions: (1) Which actors should be held liable for harm caused by AI agents? (2) What are the circumstances in which liability should arise? (3) What is the appropriate standard of care?

The first of these questions—*who* should be held liable—raises the perennial "many hands problem."[209] As with other computing technologies, the development and operation of AI agents involves multiple actors each of which may bear some responsibility for harms resulting from the technology.

---

*Toward an Automation Jurisprudence*, 6 HARV. J.L. & TECH. 103 (1992); Maruerite E. Gerstner, *Liability Issues with Artificial Intelligence Software*, 33 SANTA CLARA L. REV. 239 (1993); Curtis E.A. Karnow, *Liability for Distributed Artificial Intelligences*, 11 BERKELEY TECH. L.J. 147 (1996). This literature has grown significantly in recent years. *See, e.g.*, David C. Vladeck, *Machines Without Principals: Liability Rules and Artificial Intelligence*, 89 WASH. L. REV. 117 (2014); Paulius Čerka et al., *Liability for Damages Caused by Artificial Intelligence*, 31 COMPUT. L. & SEC. REV. 376 (2015); Peter M. Asaro, *The Liability Problem for Autonomous Artificial Agents*, PROC. 2016 AAAI SPRING SYMPOSIUM 190 (2016); Bartosz Brożek & Marek Jakubiec, *On the Legal Responsibility of Autonomous Machines*, 25 ARTIF. INTEL. & L. 293 (2017); Mark A. Chinen, *The Co-Evolution of Autonomous Machines and Legal Responsibility*, 20 VA. J.L. & TECH. 338 (2016); Ryan Abbott, *The Reasonable Computer: Disrupting the Paradigm of Tort Liability*, 86 GEO. WASH. L. REV. 1 (2018); Bathaee, *supra* note 196; Karni Chagal-Feferkorn, *The Reasonable Algorithm*, 2018 U. ILL. J.L. TECH. & POL'Y 111; Karni A. Chagal-Feferkorn, *Am I An Algorithm or a Product? When Products Liability Should Apply to Algorithmic Decision-Makers*, 30 STAN. L. & POL'Y REV. 61 (2019); JACOB TURNER, ROBOT RULES: REGULATING ARTIFICIAL INTELLIGENCE (2019); Casey, *supra* note 197; RYAN ABBOTT, THE REASONABLE ROBOT: ARTIFICIAL INTELLIGENCE AND THE LAW (2020); Andrew D. Selbst, *Negligence and AI's Human Users*, 100 B.U. L. REV. 1315 (2020); Rachum-Twaig, *supra* note 17; Lior, *supra* note 17; CHESTERMAN, *supra* note 26; Jennifer Cobbe & Jatinder Singh, *Artificial Intelligence as a Service: Legal Responsibilities, Liabilities, and Policy Challenges*, 42 COMPUT. L. & SEC. REV. 105573 (2021); Amy L. Stein, *Assuming the Risks of Artificial Intelligence*, 102 B.U. L. REV. 979 (2022); Paulo Henrique Padovan et al., *Black is the New Orange: How to Determine AI Liability*, 31 ARTIF. INTEL. & L. 133 (2022); Mihailis E. Diamantis, *Vicarious Liability for AI*, 99 IND. L.J. 317 (2023); Henderson, Hashimoto & Lemley, *supra* note 52; Anna Beckers & Gunther Teubner, *Responsibility for Algorithmic Misconduct: Unity or Fragmentation of Liability Regimes?*, 25 YALE J.L. & TECH. 76 (2023); Catherine M. Sharkey, *A Products Liability Framework for AI*, 25 COLUM. SCI. & TECH. L. REV. 240 (2024); Ayres & Balkin, *supra* note 17; Mihailis E. Diamantis, *Reasonable AI: A Negligence Standard*, 77 VAND. L. REV. (forthcoming 2025).

[209] *See* Helen Nissenbaum, *Accountability in a Computerized Society*, 2 SCI. & ENG'G ETHICS 25, 28–32 (1996), adapting the term from Dennis F. Thompson, *Moral Responsibility of Public Officials: The Problem of Many Hands*, 74 AM. POLIT. SCI. REV. 905 (1980). *See also* A. Feder Cooper et al., *Accountability in an Algorithmic Society: Relationality, Responsibility, and Robustness in Machine Learning*, PROC. 2022 ACM CONF. FAIRNESS, ACCOUNTABILITY & TRANSPARENCY 864, 867–69 (2022) (illustrating that AI systems exacerbate the many hands problem); Jennifer Cobbe et al., *Understanding Accountability in Algorithmic Supply Chains*, PROC. 2023 ACM CONF. FAIRNESS, ACCOUNTABILITY & TRANSPARENCY 1186, 1189–90, 1194–95 (2023) (suggesting that industry actors may strategically introduce the many hands problem into AI supply chains).

In the case of AI agents, these include actors that design these systems and their many constituent parts, actors that deploy AI agents and make them available to others, and actors that use AI agents for particular applications.[210] Allocating liability among these actors is a thorny challenge. If responsibility or culpability for harm were the primary criterion for allocating liability, then actors involved in the design and operation of AI agents might point to the relative autonomy of AI agents in order to absolve themselves of liability.[211] The naïve application of agency law could exacerbate the problem.

> Whether a person is liable for the torts of their agent depends on the degree of control they exercise over the agent's behaviour. The more autonomous the agent, the less likely it is that the principle [sic.] will be held liable. . . . Yet, this principle of agency law would create dangerous perverse incentives . . . operators of AI programs could avoid liability by failing to control the AI programs they operate.[212]

---

[210] *See* Ayres & Balkin, *supra* note 17, at 1; Chan et al., *Visibility*, *supra* note 14, at 962; Scherer, *supra* note 17, at 287; Lemley & Casey, *supra* note 142, at 1354–55; Rachum-Twaig, *supra* note 17, at 1171–73; Bayern, *supra* note 21, at 40–41.

[211] This is described as the "computer as scapegoat" problem. *See* Nissenbaum, *supra* note 209, at 34–35. *See also* Cooper et al., *supra* note 209, at 870 (criticizing the term "accountable algorithms" because "it makes algorithms the subject of accountability even though algorithms are not bearers of moral agency and, by extension, moral responsibility.") This concern has long been voiced by many legal scholars. *See, e.g.*, Karnow, *supra* note 208, at 189 ("The temptation to treat sophisticated intelligent agents as independent legal entities, thus absolving the humans involved, is powerful."); Jack M. Balkin, *The Three Laws of Robotics in the Age of Big Data*, 78 Ohio St. L.J. 1217, 1223 (2017) ("AI agents … are the devices through which these social relations are produced, and through which particular forms of power are processed and transformed. … the problem is not the robots; it is the humans"). Recent writing on AI agents recognizes this concern. *See, e.g.*, Chan et al., *Agentic Systems*, *supra* note 14, at 654 ("One objection against framing algorithmic systems as agents is that it distracts from the responsibility of humans."); Shavit et al., *supra* note 56, at 3 ("it is important that at least one human entity is accountable for every uncompensated direct harm caused by an agentic AI system."); Ayres & Balkin, *supra* note 17, at 2 ("people should not be able to obtain a reduced duty of care by substituting an AI agent for a human agent."). Some legal scholars, however, have shown interest in holding AI agents directly accountable, at least under certain conditions. *See, e.g.*, Wein, *supra* note 17, at 114; Chopra & White, *supra* note 17, at 145–50; Jaap Hage, *Theoretical Foundations for the Responsibility of Autonomous Agents*, 25 Artif. Intel. & L. 255, 255 (2017); Scherer, *supra* note 17, at 287. *Cf.* Bellia, *supra* note 17, at 1061–62; Chesterman, *supra* note 26, at 89–90; Bayern, *supra* note 21, at 37.

[212] Oliver, *supra* note 21, at 80–81. *See also* Chesterman, *supra* note 26, at 89–90 ("In the case of AI systems, the most difficult liability questions will arise when they operate as more than tools or instruments, beyond the control or direction of the user. In such cases, the agency relationship is actively unhelpful in that it presumes an underlying responsibility on the part of the AI system itself."); Bayern, *supra* note 21, at 37 ("agency law's significant rules about the liability of agents … would have no place in a legal regime that did not recognize the legal personhood of algorithmic agents").

An alternative, and more pragmatic, criterion for allocating liability among the actors involved in designing, operating, and using AI agents is to consider each actor's (i) *ex ante* ability to prevent harm and (ii) resources to remedy harm *ex post*. This criterion could be evaluated by assessing an actor's access to information about an AI agent (e.g., results of safety tests), its ability to alter the agent's design or operation,[213] and, more broadly, an actor's technical and financial resources that could support preventive measures or remedy harms after the fact.[214]

Turning to the second question concerning the establishment of a liability regime for AI agents—namely, determining the *circumstances* in which liability should arise—the law of agency is a more helpful guide. The Restatement (Third) of Agency holds a principal liable for an agent's tortious conduct where that conduct is "within the scope of the agent's actual authority or ratified by the principal."[215] This rule captures some of the circumstances in which a liability regime could apply to individuals or entities that use AI agents in a manner that ultimately results in harm.[216] The Restatement, however, also holds a principal liable "if the harm was caused by the principal's negligence in selecting, training, retaining, supervising, or otherwise controlling the agent."[217] This rule significantly expands the circumstances in which liability arises. Liability is not confined to the narrow scope in which an agent is deployed, but extends to decisions concerning its selection, training, and oversight. In the context of AI agents, a comparable rule could be established to impose liability on entities engaged in designing, deploying, and supervising AI agents. Finally, the Restatement of Agency includes an important qualification: liability is limited to harms that are foreseeable.[218] While generally justified on moral and economic grounds, it

---

[213] *See, e.g.*, EU AI Act, *supra* note 158, at art. 25 (imposing obligations on actors who make "substantial modifications" to certain AI systems). For an incisive analysis concerning the allocation of liability that pre-dates the EU AI Act, see Scherer, *supra* note 17, at 288. *See also* Cobbe et al., *supra* note 209.

[214] *See* Buiten, *Product Liability for Defective AI*, *supra* note 208, at 18–19 (arguing that because neither developers nor users have complete control over autonomous systems both should share liability.) For further discussion on the allocation of liability, see Buiten et al., *The Law and Economics of AI Liability*, *supra* note 208, at 11–13.

[215] RESTATEMENT (THIRD) OF AGENCY § 7.04 (2006).

[216] For the avoidance of doubt, as noted at the outset, the analysis in this Article is not concerned with the direct legal application of agency law to AI agents.

[217] *Id.* at § 7.05(1).

[218] *Id.* at § 7.05 cmt. d ("Conduct that results in harm to a third person is not negligent or reckless unless there is a foreseeable likelihood that harm will result from the conduct."). *See also* Dalley *supra* note 20, at 497 ("The foundational principle of agency law is that the principal, who has chosen to conduct her business through an agent, must bear the *foreseeable* consequences created by that choice") (emphasis added). *See id.* at 501 ("The

is unclear whether this foreseeability requirement is appropriate for AI agents whose behavior is highly unpredictable and, at least for the time being, subject to only limited visibility and accountability.[219]

The third question regarding the establishment of a liability regime for AI agents concerns the appropriate *standard of care*. Here too agency law offers useful guidance. According to the Restatement of Agency, an agent must "act with the care, competence, and diligence normally exercised by agents in similar circumstances", which is informed by "[s]pecial skills or knowledge possessed by an agent".[220] Reflecting on this rule in the context of AI agents highlights important issues. For example, it suggests that AI agents developed or adapted for a specific domain should be held to a higher standard (compared with general-purpose AI agents operating in that same domain). This rule, at first glance, seems reasonable. The more capable an AI agent in a given domain, the more we ought to expect from it in that domain. The problem with this rule is that it could incentivize well-resourced AI developers, such as OpenAI and Google, to refrain from building domain-specific agents, lest they be held to a higher standard of care and exposed to liability. The development and operation of domain-specific agents might then be left to other, less well-resourced companies that ultimately produce less reliable and less safe AI agents.

CONCLUSION

The governance of AI agents presents challenging tradeoffs. Navigating these tradeoffs will become increasingly consequential as AI agents are used more widely and are entrusted to perform more complex and sensitive tasks. As illustrated in the economic theory and law of agency relationships, the greater the opportunities in delegating work to an agent, the greater the associated risks. The enduring problems of information asymmetry, authority, and loyalty are now beginning to emerge in a new context, shaped by the distinct features of AI agents. While traditional mechanisms for tackling principal-agent problems are instructive, developing tools for incentive design, monitoring, and enforcement for AI agents remains a

---

principal's responsibility is limited to foreseeable consequences consistent with both moral and economic reasoning.") For further discussion on foreseeability and harms arising from AI, see Karnow, *supra* note 208, at 178–81; Weston Kowert, *The Foreseeability of Human–Artificial Intelligence Interactions*, 96 TEX. L. REV. 181 (2017); Bathaee, *supra* note 196, at 922–24; Selbst, *supra* note 208, at 1331–46; Ayres & Balkin, *supra* note 17, at 2.

[219] *See* Lemley & Casey, *supra* note 142, at 1357–58 ("To efficiently deter behavior, we must be able to predict it. But if we don't know how the robot will behave because it might discover novel ways of achieving the goals we specify, simply pricing in the cost of bad outcomes might have unpredictable effects.").

[220] RESTATEMENT (THIRD) OF AGENCY § 8.08 (2006).

formidable challenge. New governance principles are needed. These should center around expanding the range of interests that AI agents serve, improving visibility into the design and operation of these agents, and holding developers, deployers, and users accountable if and when harm occurs. Principles, however, are not enough. The effective governance of AI agents also requires new technical and legal infrastructure. Given the technology is still in its infancy, policymakers and companies building AI agents have a window of opportunity. They should take it, and soon.